# Evaluating the effectiveness, reliability and efficiency of a multi-objective sequential optimization approach for building performance design

**Riccardo Talami [a] [b] *, Jonathan Wright [c], Bianca Howard [d]**

[a] Department of Architecture, College of Design and Engineering, National University of Singapore, 117566, Singapore.

[b] Singapore-ETH Centre, Future Resilient Systems II, 1 Create Way, 138602, Singapore.

[c] School of Architecture, Building and Civil Engineering, Loughborough University, Loughborough, United Kingdom.

[d] Department of Mechanical Engineering, Columbia University, 220 S.W. Mudd Building 500 W 120th St, New York, NY, USA.

E-mail (Corresponding author): rtalami@nus.edu.sg

## ABSTRACT

The complexity of performance-based building design stems from the evaluation of numerous candidate design options, driven by the plethora of variables, objectives, and constraints inherent in multi-disciplinary projects. This necessitates optimization approaches to support the identification of well performing designs while reducing the computational time of performance evaluation. In response, this paper proposes and evaluates a sequential approach for multi-objective design optimization of building geometry, fabric, HVAC system and controls for building performance. This approach involves sequential optimizations with optimal solutions from previous stages passed to the next. The performance of the sequential approach is benchmarked against a full factorial search, assessing its effectiveness in finding global optima, solution quality, reliability to scale and variations of problem formulations, and computational efficiency compared to the NSGA-II algorithm. 24 configurations of the sequential approach are tested on a multi-scale case study, simulating 874 to 4,147,200 design options for an office building, aiming to minimize energy demand while maintaining thermal comfort. A two-stage sequential process-(building geometry + fabric) and (HVAC system + controls) identified the same Pareto-optimal solutions as the full factorial search across all four scales and variations of problem formulations, demonstrating 100% effectiveness and reliability. This approach required 100,700 function evaluations, representing a 91.2% reduction in computational effort compared to the full factorial search. In contrast, NSGA-II achieved only 73.5% of the global optima with the same number of function evaluations. This research indicates that a sequential optimization approach is a highly efficient and robust alternative to the standard NSGA-II algorithm.

# 1. Introduction

## 1.1 Background

The design of buildings based on their performance is increasingly becoming a key practice in efforts to reduce energy consumption and decarbonize the built environment [1]. This involves evaluating and selecting design solutions from a number of candidate design options to address design objectives while adhering to building regulations [2]. The increasing adoption of Building Performance Simulation (BPS) tools has facilitated the performance assessment of design options [3]. Nevertheless, the iterative trial-and-error methods often employed for selecting design solutions can be time-consuming and may fail to consider potential design alternatives [2], [4]. In this context, computational methods for design optimization offer great potential in supporting the identification of solutions that meet design objectives while accounting for design constraints [5]. This approach involves the integration of numerical simulation through BPS software with mathematical optimization, in the form of optimization algorithms [6]. Automating this process transforms the time-consuming and often ineffective task of performance-based building design into a rapid evaluation of design options. This enables a more comprehensive exploration of the solution space, leading to the identification of well performing design solutions.

Performance-based building design encompasses architectural elements (building geometries and fabric), HVAC system design, and operational controls, necessitating the engagement of various stakeholders, such as architects, engineers, building managers and owners. However, this represents a complex task of design exploration which relies on multi-disciplinary knowledge with interacting and inter-dependent factors, often characterized by conflicting and highly constrained design objectives [7]. In addition, this results in the need to evaluate a large number of candidate design options stemming from the plethora of variables, objectives and constraints which characterize the multi-dimensionality of real-world problems [8]. These considerations reveal the need to integrate optimization approaches into performance-based building design to facilitate the identification of well performing design options while overcoming the computational burden of performance evaluation.

## 1.2 Contribution of this paper

In response, this paper proposes and evaluates a sequential approach for multi-objective design optimization of building geometry, fabric, HVAC system and controls, advancing performance-based building design. Unlike conventional multi-objective optimization approaches that require exploration of all variables simultaneously, the sequential approach integrates each building element (building geometry, fabric, HVAC system and controls) in a cohesive sequence, leveraging their synergies and interdependencies. Each stage identifies Pareto-optimal solutions, passing only the optimal results to the next stage while discarding sub-optimal ones. This systematic process reduces computational demands, streamlines exploration and evaluation, and simplifies integration into design workflows. By addressing practical constraints such as resource and time limitations, the approach lowers technical barriers and promotes wider adoption of performance-driven strategies by building stakeholders.

A benchmarking framework was developed to evaluate the sequential design optimization approach against a full factorial search in terms of its ability to find global optima, computational performance, and solution quality. The reliability of the sequential search was tested across four problem scales—very small (874 design options), small (52,400), medium (345,600), and large (1,036,800)—and four variants of the large scale (4,147,200 options), incorporating combinations of internal load (low/high) and weather conditions (cold/hot years). Its efficiency was also compared to the Non-Dominated Sorting Genetic Algorithm II (NSGA-II), a widely used population-based metaheuristic. Twenty-four configurations of the sequential search were tested, varying initial starting points (low, middle, upper, and random bounds), grouping strategies (no grouping, grouping by design element, and grouping by field), and initialization settings (initial and iterative). A multi-scale case study of an open-office building involving 11 design variables—spanning geometry, fabric, HVAC systems, and controls—has been simulated to generate an exhaustive dataset of 874 to 4,147,200 design options. Optimization objectives focused on minimizing heating energy demand and occupied hours outside the comfort zone.

This paper is organized as follows. Section 2 reviews existing studies in the literature that evaluated multi-stage optimization approaches for building design. Section 3 details the sequential design optimization approach and algorithm proposed in this paper. In Section 4, the methodological workflow to evaluate the performance of the sequential design optimization is highlighted with details of the evaluation framework and experimental tests, application to a multi-scale case study, and modelling and simulation workflow. Section 5 analyses the results of the evaluation of the proposed optimization approach while Section 6 discusses the findings. Finally, the main conclusions are summarized in Section 7.

## 2. Literature review

Previous studies have optimized building design elements through multiple sequential stage-dependent processes, with each stage building on prior results. This represents a shit from conventional multi-objective optimization which involves the simultaneous exploration of the search space across all variables. Table 1 benchmarks previous multi-stage optimization studies with the proposed approach, highlighting differences in methodologies, objectives, and outcomes.

Table 1. Literature matrix overview of sequential multi-stage optimization approaches for building performance design.

| Authors and Year | Objectives | Building Design Elements | Optimization approach | Climate context | Key findings |
|---|---|---|---|---|---|
| Verbeeck et al., 2007 | Minimize heating demand, environmental impact, investment costs. | Fabric, HVAC system | Two-stage with GA | Belgium | Decoupling envelope and HVAC optimization improves performance compared to reference case |
| Hamdy et al., 2013 | Minimize operational costs and energy consumption | Fabric, HVAC system, Renewables | Three-stage with controlled elitist GA variant | Finland | Multi-stage optimization reduces computational costs while finding effective solutions for energy and environmental objectives |
| Ascione et al., 2019 | Minimize energy consumption, lifecycle global cost, and investment cost | Geometry, Fabric, HVAC systems, Renewables | Three-phase GA and exhaustive search | Milan, Italy | Multi-objective approach improved energy performance, lifecycle costs, and investment costs compared to a reference case |
| Evins, 2015 | Minimize carbon emissions and investment costs | Geometry, HVAC system, Controls | Multi-level Optimization with GA and mixed-integer linear programming | London, UK | Layered optimization of design and operational variables led to efficient integration of building and system performance |
| This Study | Minimize heating demand and discomfort hours | Geometry, Fabric, HVAC system, Controls | Sequential Optimization with exhaustive search at each stage | Nottingham, UK | Sequential optimization is effective, reliable and efficient, accounts for dependencies between elements |

Verbeeck et al. [9] utilized a genetic algorithm (GA) to optimize the heating demand, environmental impact, and investment costs of a dwelling in Belgium. The optimization was carried out in two steps. Initially, they optimized envelope-related variables with respect to the three objectives. Subsequently, they combined the 44 optimal envelope solutions with system-related measures. Their findings suggested that because decisions regarding the building envelope have a longer-lasting impact on energy performance compared to most heating system components which have different lifespans, it is preferable to decouple the optimization of architectural and engineering design elements. Similarly, Hamdy et al. [10] developed a multi-stage optimization approach to identify cost-optimal and nearly net-zero energy building solutions for a single-family house case study in Finland. They employed an algorithm that combines deterministic methods with a controlled elitist GA, a variant of the NSGA-II. In the first stage, the algorithm identified the optimal combinations of building envelope and heat recovery variables. Stage 2 assessed the financial and environmental viability of the cost-optimal building designs from the first stage by combining each optimal solution's space heating energy demand with the HVAC system to obtain new optimal designs. Finally, in stage 3, the economic and environmental viability was further improved by adding renewable energy systems. This method resulted in a reduced computational effort while extensively exploring the solution space. Ascione et al. [11] developed a framework for multi-phase, multi-objective design optimization to enhance building performance. The process was divided into three sequential phases to optimize building geometry, opaque and transparent envelope, HVAC systems, and their operation. During phase 1, A GA was used to optimize building envelope, geometry, and control setpoints. Subsequently, an exhaustive sampling was conducted on design alternatives to identify the optimal energy system configuration and renewable energy systems, based on the loads determined in Phase 1. Finally, in phase 3, a multi-criteria decision-making approach was applied to determine the optimal solution. The framework was tested on a typical office building in Milan, aiming to minimize energy consumption, lifecycle global cost, and investment cost, compared to a reference base case. The multi-stage design optimization process was found to deliver better performance compared to the reference design. Evins [12] proposed a multi-level optimization framework for building design, optimizing building-related variables and energy supply system design (upper level) before addressing operational variables (lower level). The optimization of operational variables depended on the results from the upper level. Initially, a genetic algorithm (GA) was used to find optimal solutions for building and system design, focusing on minimizing carbon emissions and investment costs. This step also determined the building's energy demand. During the second step, the operational variables were optimized using mixed-integer linear programming (MILP). The study concluded that the multi-level optimization approach effectively exploits synergies between different building design elements, resulting in more efficient overall solutions.

While the literature validates multi-stage approaches for building design optimization, significant gaps remain. Most studies assess optimal solutions by comparing them to a baseline design, offering limited insights into the overall effectiveness, reliability, and efficiency of the methods. Furthermore, although multi-stage optimization reduces computational time for evaluating numerous solutions, it often neglects how synergies and dependencies between building elements are maintained.

Previous research considers multiple design elements but rarely addresses all four critical components—building geometry, fabric, HVAC systems, and controls—essential for performance-based design. This limits the applicability and integration of these methods into the design process. Additionally, commonly employed metaheuristic algorithms, such as evolutionary and particle swarm algorithms [13], [14], rely on probabilistic search operators that require meticulous parameter tuning for convergence [15]. Their variability across repeated runs complicates decision-making [16], as reliable identification of the global optima is critical. The complexity of these approaches further hampers adoption by stakeholders who may lack the expertise or training to implement them effectively.

These gaps highlight the need for optimization methods that capture synergies and interdependencies between design elements while reducing algorithmic complexity and computational effort. Such approaches would better integrate into the design process and encourage broader stakeholder adoption. To address these challenges, this paper proposes a novel sequential approach for multi-objective optimization of building geometry, fabric, HVAC systems, and controls. The method effectively captures design synergies, offering a practical, reliable, and efficient solution for performance-based building design.

## 3. The sequential design optimization approach

To support full reproducibility of the sequential optimization approach, this paper provides a comprehensive step-by-step breakdown of each optimization stage, detailing the mathematical foundation and pseudocode to guide implementation. Each computational step in the pseudocode serves as a practical roadmap, enabling researchers and practitioners to adapt the approach to project-specific requirements. The flexibility of this approach allows users to adjust design variables and objectives without changing the foundational logic of the optimization stages, facilitating its application across various building designs and real-world implementations.

As depicted in Figure 1, this approach involves conducting individual design optimizations at each step of the process to identify a set of Pareto optimal solutions. Each optimization occurs sequentially, with only optimal solutions from previous stages being passed to the next stage, while sub-optimal solutions are discarded. Design options at each stage are optimized while concurrently fixing the next stages on a baseline scenario. This fixed baseline setup ensures that interdependencies among variables are always inherently considered, as every optimization stage evaluates the effects of changing one subset of variables against a consistent backdrop of the other design elements. By carrying forward the optimal solutions from each stage, this approach ensures that all design decisions are made with respect to the full system of building parameters, capturing complex synergies that arise from the combined effects of building design elements. The systematic transfer of optimal configurations ensures each stage builds on well-performing solutions, reducing the risk of unsuitable or random pairings across design variables. This iterative refinement maintains consistency, reliably identifying optimal multi-variable configurations while effectively capturing and leveraging interdependencies among design variables.

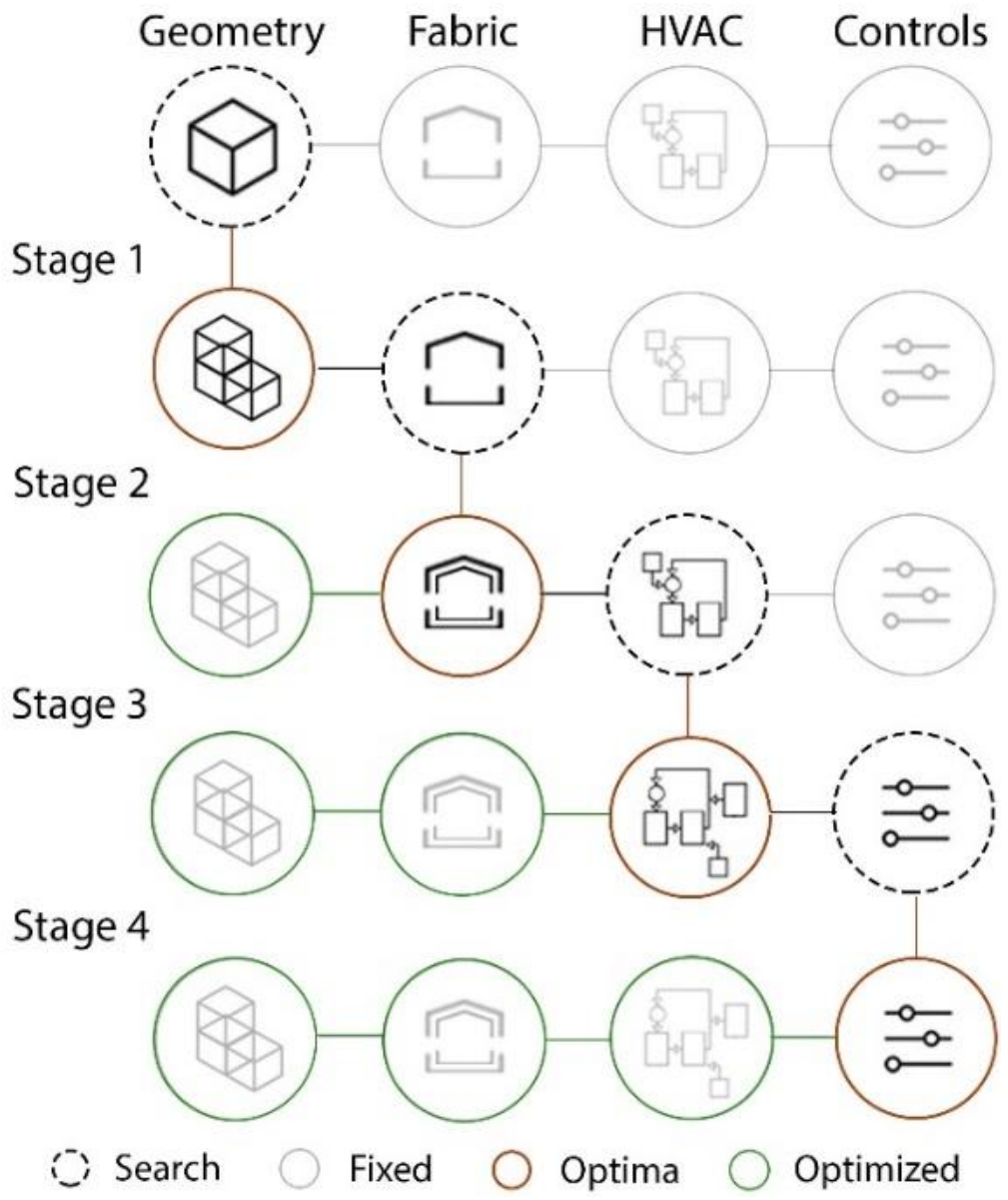

Figure 1. Schematic diagram of the sequential design optimization approach.

Furthermore, optimizations at each stage are performed using a full factorial (exhaustive) search to ensure consistent convergence to the global optimal set of solutions. An exhaustive search, which does not require tuning, guarantees the identification of the global set of optimal solutions for the defined problem across repeated runs of the search [17]. Additionally, this method is independent of the number of objective and constraint functions, making it well-suited for developing an optimization algorithm capable of comprehensively assessing the numerous variables, objectives, and constraints inherent in the multi-disciplinary nature of real-world building design problems. The extensive computational effort associated with the exhaustive search - due to the evaluation of all possible solutions - is mitigated by sequential steps that narrow down the search space at each stage. This method replicates the integrated interdisciplinary decision-making process expected in practical design scenarios, where interactions among variables are essential for an optimized, holistic design solution.

The steps of the sequential optimization approach can be mathematically expressed as:

1) **Optimization Stages and Variables**: Let $x_z$ represent a solution at stage $z$, and $X_z$ the set of design solutions at stage $z$. $f(x_z)$ represents a single objective function at stage $z$, while if the optimization is multi-objective, the objective function is assumed to be a vector $F(x_z) = (f_1(x_z), f_2(x_z),\ldots,f_m(x_z))$ where $m$ is the number of objectives.
2) **Fixing Future Stages to a Baseline**: For each stage $z$, the variables in future stages $j > z$ are set to baseline values $x_j^{baseline}$ to ensure that the optimization accounts for interdependencies and synergies between design elements (1):

$$For\ j > z,\ x_j = x_j^{baseline} \tag{1}$$

3) **Sequential Optimization**: At each stage $z$, an exhaustive search (full factorial search) is conducted over all possible combinations in the set $X_z$. The set of Pareto-optimal solutions at stage $z$ is denoted as $P_z \subseteq X_z$. These are the non-dominated solutions from the full factorial search conducted at that stage (2):

$$P_z = \{x_z \mid x_z \in Factorial\ Search\ (X_i), x_{z-1} \in P_{z-1}\} \quad (2)$$

The sequential nature implies that the set of solutions passed from one stage to the next includes only the Pareto-optimal set from the previous stage. At each stage $z$, only the Pareto-optimal solutions from stage $z-1$, denoted as $P_{z-1}$, are considered for further optimization. Therefore, for each stage $z$, the sequential optimization can be written as:

$$x_z = \min_{x_z \in X_z} F(x_z) \quad subject\ to\ x_{z-1} \in P_{z-1}, x_j = x_j^{baseline} for\ j > z \quad (3)$$

## 3.1 The sequential design optimization algorithm

The initialization of the algorithm encompasses two settings: the initial starting bound and the grouping category. The initial starting bound establishes the fixed baseline values of design parameters at the commencement of the search process, serving as proxy for identifying a starting point within the search space. The initial starting bound can be determined by rules of thumb and the experience of the design team or selected randomly. The grouping category allows to categorize the design variables within certain groups. This determines (1) the number of sequential design steps in the search process and (2) the size of the search space at each step. By default, the optimization of each variable represents an individual step in the sequence. However, grouping of variables that are optimized concurrently reduces the number of optimization steps. Additionally, given the use of a full factorial search at each step, the number of solutions explored is determined by the included variables and their perturbation values.

Upon algorithm initialization, the set of solutions from the first stage of the search, is generated, with subsequent stages fixed based on their initialized baseline value. The search evaluates the objective function of solutions in the first stage, exhaustively exploring their perturbation values. Suboptimal solutions are discarded, and Pareto optimal solutions are extracted and inserted in the next stage of the sequential search. This process generates a new set of solutions containing both new and optimal solutions from the previous stage, thereby restricting the search space and allowing only Pareto optimal solutions to advance to the next stage. The loop, involving solutions generation, objective function evaluation, Pareto optimal solution extraction, and insertion into a new set of solutions, is iteratively performed until all sequential steps, as determined by the initialization process through the grouping category, are completed. The final Pareto optimal solutions are then returned for assessment by decision-makers.

Depending on the scope of design optimization and factors such as available computational run-time, decision-makers can choose to terminate the process or continue the design optimization through iterative runs of sequential searches. In the iterative run, Pareto optimal solutions from the initial run serve as starting points for the sequential search. Thus, while the initial run starts with a single point, the iterative begins with as many starting points as the number of optima identified in the initial run. This approach allows the search to start from regions of the search space likely containing the global optima, unless they already match the optimal solutions from the initial search. The iterative run begins with the first optimal solution from the initial run and initializes the sequential process based on the variable values. The loop, including solutions generations, objective function evaluation, Pareto optimal solution extraction, and insertion into a new set of solutions, is performed iteratively until all sequential steps are completed. The Pareto optimal solutions are then returned and stored. The algorithm then proceeds to the next optimal solution from the initial run, initiating the sequential optimization process. The loop, involving the storage of optima from each sequential optimization and transitioning to the next optimal solution from the initial run, is iteratively carried out until all optimal solutions are sequentially optimized. Solutions stored at each iteration form the final population are then evaluated. The resulting final Pareto optimal solutions can finally be assessed by the decision-makers.

```
# Initialization

SET initial_bounds
SET grouping_category (z stages)
SET solutions_set = []

# Initial Run

FOR each z stage IN grouping_category DO
    IF solutions_set IS EMPTY THEN
        GENERATE solutions_set USING initial_combinations
    ELSE
        GENERATE next_solutions_set COMBINING solutions_set WITH Pareto-optimal solutions FROM previous stages
        SET solutions_set = next_solutions_set
    END IF
    EVALUATE objective f(x) FOR each solution x IN solutions_set
    EXTRACT Pareto-optimal solutions BASED ON f(x)
    SET solutions_set = Pareto-optimal solutions
END FOR

RETURN final_solutions_set AS Pareto-optimal solutions

# Iterative Run

SET Pareto-optimal solutions FROM final_solutions_set

WHILE NOT termination_criteria_met DO
  FOR each solution x IN Pareto-optimal solutions DO
    INITIALIZE solutions_set FOR solution x
    FOR each z stage IN grouping_category DO
        IF solutions_set IS EMPTY THEN
            GENERATE solutions_set USING initial_combinations
```

```
            ELSE
                GENERATE next_solutions_set COMBINING solutions_set WITH Pareto-optimal solutions FROM previous stages
                SET solutions_set = next_solutions_set
            END IF
            EVALUATE objective f(x) FOR each solution x IN solutions_set
            EXTRACT Pareto-optimal solutions BASED ON f(x)
            SET solutions_set = Pareto-optimal solutions
        END FOR
        SWITCH TO next_Pareto_optimal_solution
    END FOR
    CHECK termination_criteria: IF all solutions have been evaluated THEN TERMINATE
END WHILE

RETURN final_solutions_set AS Pareto-optimal solutions
```

# 4. Methodology

The methodological workflow to evaluate the performance of the sequential design optimization approach consist of the development of 1) a framework encompassing various evaluation criteria (Section 4.1), 2) a case study highlighting a number of building design variables, objectives and constraint functions (Section 4.2), 3) a set of experimental tests to conduct (Section 4.3), and 4) a modelling and simulation workflow (Section 4.4).

## 4.1 Evaluation framework

A framework for evaluating the performance of sequential design optimization has been developed (Figure 2). This framework relies on the benchmarking of the sequential search approach against a full factorial search. This is due to the ability of the full factorial search, which, for the defined problem, is guaranteed to identify the global set of optimal solutions. Note that the outcomes derived from this comparison are intended to shed light on the assessment of the approach, rather than guiding the design process by identifying the best-performing solutions. Additionally, the reliability of the sequential search is evaluated across four scales of problem formulation: very small (874 design options), small (52,400 design options), medium (345,600 design options), and large (1,036,800 design options). The reliability is further tested on four variants of the large-scale problem (4,147,200 design options), generated by varying two input boundary conditions: internal loads (low and high) and weather conditions (cold and hot years). Finally, the efficiency of the sequential search is compared to a widely used population-based metaheuristic search algorithm, the Non-Dominated Sorting Genetic Algorithm (NSGA – II). The evaluation framework includes the following criteria and method of analysis.

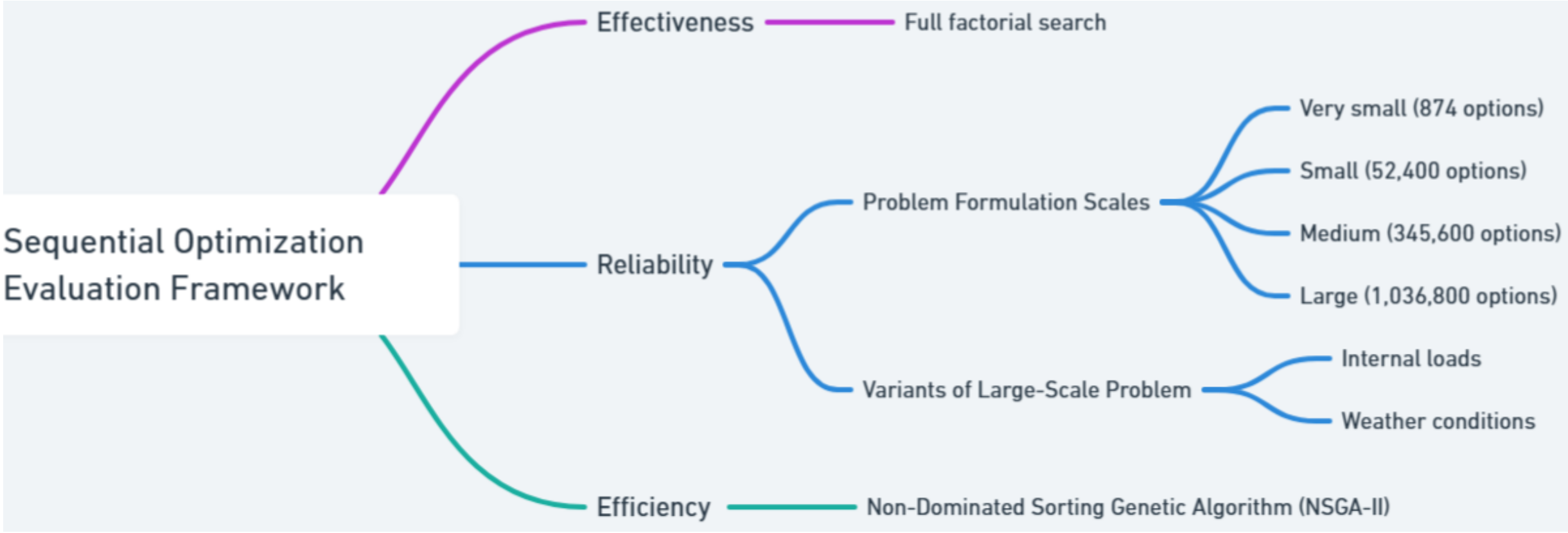

Figure 2. Evaluation framework developed to assess the performance of the sequential optimization approach.

### 4.1.1 Search effectiveness in finding global optimal solutions

The effectiveness of the search is defined as its ability to identify the global set of optimal solutions. In this paper, the Pareto optimal solutions are classified into two categories: global and search optima. The term "global" optima refers to the solutions identified by the full factorial search. The term “search” optima denotes instead the solutions found by the sequential search that do not overlap with the global optima. As an example, assuming that the sequential search identified 15 optimal solutions: all 15 can be global optima (same solutions identified by the full factorial), none are global optima (different solutions found by the full factorial), or a portion of them are global optima (out of 15, 5 are the same solutions identified by the full factorial search, and therefore global optima, and 10 are different solutions obtained by the full factorial, and therefore “search” optima).

The magnitude of effectiveness is calculated as the ratio between the number of global optima identified by the sequential search and the total number of global optima found by the full factorial search (Equation 4):

$$Effectiveness = \frac{Ng^{sequential\ search}}{Ng^{full\ factorial}} \tag{4}$$

Where $N_g$ refers to the number of global optimal solutions.

### 4.1.2 Computational performance

The computational performance refers to the computational load of the sequential search and the computational savings with respect to the full factorial search. The computational load refers to the number of function evaluations of the sequential search to identify the Pareto optimal solutions. The computational savings are the savings in terms of number of function evaluations with respect to the full factorial search. Utilizing the number of function evaluations as a metric to quantify the computational performance of the search yields advantages over commonly utilised methods based instead on the calculation of the computational time. In fact, the computational time might vary across multiple runs, it is dependent on the characteristics of the machine utilised (e.g., number of cores, central processing unit – CPU), and sensitive to the adoption of parallel computing approaches. The number of function evaluations instead provides an unbiased metric, insensitive to the number of runs (for deterministic search methods) and to the characteristics of the machine utilised. Therefore, the quantification of the number of function evaluations allows for an unbiased comparison between computational performance of different algorithms. The computational load is calculated as the ratio between the number of function evaluations from the sequential search and from the full factorial search (Equation 5):

$$Computational\ load = \frac{Ne^{sequential\ search}}{Ne^{full\ factorial}} \tag{5}$$

Where $N_e$ refers to the number of function evaluations.

### 4.1.3 Solution optimality

Solution optimality is assessed through the analysis of the search and global optimal set of solutions. This is evaluated on the variables space and on the objective space. The analysis on the variable space refers to the trends of design options identified between the search and global optima. The analysis on the objective space refers to the "performance difference" of the optimization objectives between search and global optima, calculated as per Equation 6:

$$Performance\ difference = f(x_s) - f(x_g) \tag{6}$$

Where $f(x_s)$ and $f(x_g)$ refer to the objective value of the search and global optimal solution, respectively.

As depicted in Figure 3, the objective values of each optimal solution of the sequential search (identified as optimal trade-offs between energy consumption and thermal comfort) are benchmarked against the values obtained by the nearest optimum on the Pareto frontier of the full factorial search, this being the solution with the least Euclidean distance in the variable domain. This evaluation metric yields advantages over alternative methods for the quantification of the "performance difference" since it allows for a detailed assessment of each Pareto optimal solutions.

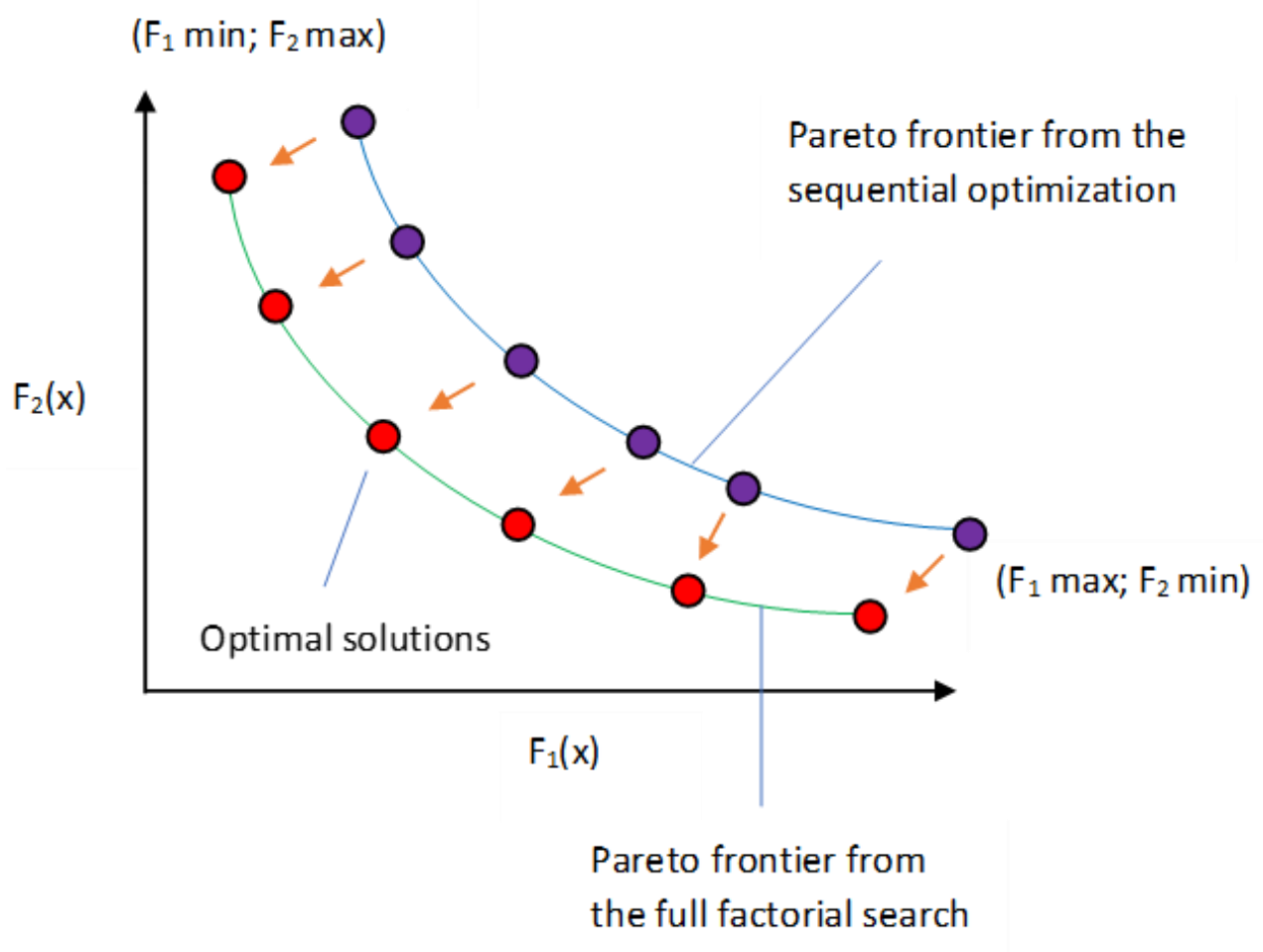

Figure 3. Performance difference between Pareto optimal solutions of the sequential optimization (purple dots) and from the full factorial search (red dots).

## 4.2 Evolutionary algorithm (NSGA-II)

The efficiency of the sequential optimization approach is evaluated as a comparison of performance against a population-based metaheuristic search algorithm, the Non-Dominated Sorting Genetic Algorithm (NSGA – II) [42]. It was chosen among other evolutionary algorithms, since it is considered one of the most used in the building performance optimization domain and it is well suited to solve multi-objective optimization problems [43]. NSGA-II operates by generating a population of solutions that are evaluated based on multiple objectives, organizing them into ranks known as Pareto fronts through a process called non-dominated sorting. Each front contains solutions that are non-dominated with respect to each other. NSGA-II then applies crowding distance calculations within each front to maintain diversity, ensuring a spread of solutions across the objective space. This structure enables NSGA-II to maintain a diverse set of high-quality solutions while optimizing for multiple objectives simultaneously. A characteristic of evolutionary algorithms is that they use probabilistic search operators (selection, crossover, and mutation), the parameters of which require careful tuning if consistent convergence onto the optima is to be obtained [15]. Selection, crossover, and mutation operations evolve the solution population. Selection prioritizes solutions in lower-ranked Pareto fronts with higher crowding distances, while crossover and mutation introduce variability, enabling the exploration of new solutions. The selection and tuning of the operators of a GA represents a trade-off between convergence reliability maintaining the exploratory power of the algorithm (exploration) while ensuring convergence velocity (exploitation) [15] [44]. Table 2 breakdowns the tuning parameters adopted.

Table 2. NSGA-II tuning parameters for performance comparison.

| GA Operator | Selected Option |
|---|---|
| Population size | 30 |
| Sampling | Same initial starting point as sequential search |
| Selection | Binary tournament (2 individuals) |
| Crossover | Uniform with a probability of 0.5 |
| Mutation | Bitflip with a probability of 0.1 |
| Termination | Same number of function evaluation as sequential search |

Since the probabilistic nature of GAs results in different solutions being found across repeated runs of the search [16], the optimization through NSGA-II was run 20 times and the four runs that obtained the highest performance were used for the comparison with the experimental tests of the sequential search four different initial starting points, that is to say starting the search process with the variable values set at their low, middle, upper, and random bounds.

## 4.3 Case study

The case study building is a ground-level 300 $m^2$ open-plan office building with a floor-to-ceiling height of 2.7m, nominally located in Nottingham (UK). A perimeter core zoning layout subdivides the indoor area into a core zone and four around the perimeter. Operable ribbon windows on the façades are modelled without any internal or external shades allowing free cooling through natural ventilation during the summer period. Tables A.1 and A.2 in Appendix A detail the values and schedules of occupancy and light density, electric equipment load, infiltration rate, outdoor air flow, and activity level applied to the case study building, based on the National Calculation Method (NCM) for office spaces [18].

### 4.3.1 Objective functions

In the context of building design exploration and decision-making, the literature predominantly focuses on three key performance indicators: energy use, thermal comfort, and cost [13], [19], [20], [21]. Consequently, this paper evaluates the performance of candidate solutions based on two objectives that require minimization: a) energy demand, and b) occupied hours outside the comfort zone. The total heating energy demand, expressed in kWh/year, represents the combined energy consumption of HVAC components (supply and distribution) needed to meet the space heating energy requirements during occupied periods. Cooling was not considered since in this climate is minimal and typically not designed for. Total discomfort hours are defined as the total occupied hours during which the Predicted Mean Vote (PMV) index, based on Fanger's comfort model [22], exceeds 0.5 or falls below -0.5. These thermal comfort thresholds align with the acceptable conditions recommended by the ASHRAE Standard 55-2017 [23] for mechanically conditioned indoor environments. The optimization goal is to minimize both the space heating energy demand during occupied periods and the total number of occupied hours that fall outside the defined comfort zone. In this paper, the constraints on the objective functions are defined through the problem formulation.

### 4.3.2 Design variables and option values

The design variables related to each building design element (building geometry, fabric, HVAC system and controls) and corresponding option values explored across the four scales of problem formulations (very small, small, medium, and large) are detailed in Table A.3 in Appendix A. These variables are considered influential design parameters based on literature and were selected due to their significant impact on the trade-off between design objectives. The selection of design variables is guided by established engineering standards, including the ASHRAE Handbook on HVAC Systems and Equipment [24], the REHVA guidebook [25], and recognized design references such as Neufert [26]. Additionally, the design option values for these variables are established in accordance with relevant building regulations and guidelines, such as the Approved Document L2A of the UK Building Regulations for new non-residential buildings [27], the Passivhaus Standard [28] , and CIBSE Guide A [29].

Three shape types were chosen to represent varying levels of complexity in form finding at the beginning of the design process: a conventional Euclidean shape (rectangular), a complex Euclidean shape (L-shaped) and a free-form shape (Figure 4). All three geometries have equal floor area, volume, and envelope wall area (compactness ratio: 0.93 1/m). This helps to isolate the performance impact of the geometrical properties of each shape, rather than the effects of heat gains and losses due to variations in envelope area.

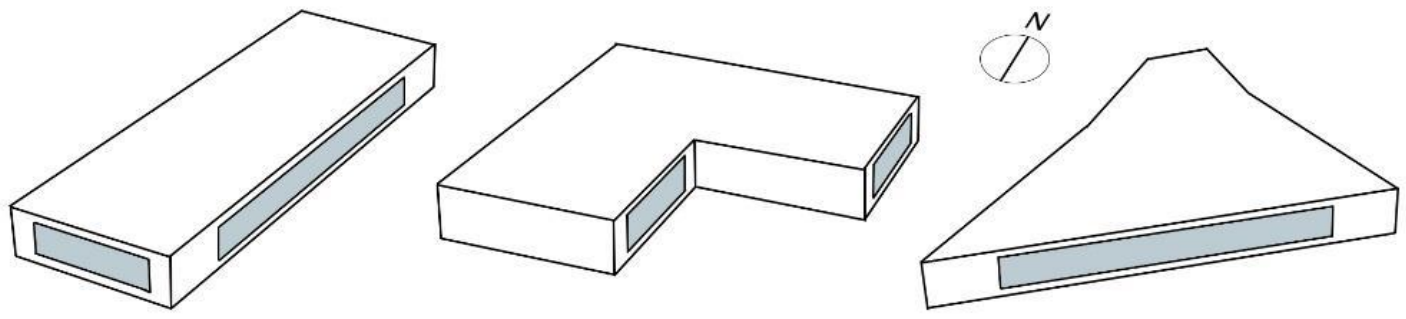

Figure 4. Building shapes and their positioning with respect to the North axis.

The window-to-wall ratio (WWR) ranges from a predominantly opaque façade with a 25% WWR to a fully glazed façade with a 95% WWR, with an additional 5% accounted for by mullions areas. Incremental WWR values of 25% are evaluated, including intermediate options at 50% and 75% WWR. The influence of building orientation on heat gains and losses is assessed by rotating each building shape by 45-degree increments from the baseline orientation (0°). Thus, four orientations are considered, ranging from 0° to 135°, to avoid redundant values for symmetrical (Euclidean) shapes. Regarding the building fabric, two constructions with similar U-values—heavyweight (HW) concrete and lightweight (LW) timber—are selected to assess the impact of thermal mass on optimal performance. Table 3 provides the U-values of the building assemblies of a lightweight and heavyweight construction. Additional details regarding material layers and thermophysical properties for each building fabric element are available in Tables A.4-A.7 in Appendix A.

Table 3. U-values of building construction options of the open-plan office.

| | **U-value (W/m$^2$K)** | |
|---|---|---|
| | *LW* | *HW* |
| Walls | 0.394 | 0.360 |
| Roof | 0.262 | 0.253 |
| Floor | 0.253 | |

Five insulation thickness options are considered. Beginning with the baseline thickness of 8.5 cm, which meets the Approved Document L2A of the UK Building Regulations for new buildings other than dwellings [27], each subsequent option increases the insulation thickness by 25%, culminating in a maximum thickness of 17 cm, or twice the initial thickness. Three types of windows are analysed as design variables, each characterized by decreasing U-values influenced by the thermal properties of the glazing, the number of panes, and the gas filling between the panes. The baseline scenario, aligned with the Approved Document L2A [27], employs double clear glazing with air (U-value of 2.8 W/m$^2$ K). A mid-range scenario is represented by triple clear glazing with air (U-value of 2 $W/m^2$ K), while the upper limit scenario features triple glazing with low emissivity coating and argon gas (U-value of 1.23 $W/m^2$ K). Additionally, two distribution systems and two generation plant types are considered for HVAC system. The distribution systems include a radiant system (radiant ceiling panel), which supplies over 50% of the total space heating load through thermal radiation, and a conventional 'all-air' system (VAV), which primarily operates via convection. The generation plant options comprise a conventional boiler and a heat pump. Five supply water temperatures are selected based on the ASHRAE handbook [30], with a nominal median value of 40°C. The range includes lower and upper bounds of 30°C and 50°C, with incremental steps of a 5°C (35°C and 45°C). For HVAC controls, two variables are selected: heating setpoint and setback air temperatures. Six values for each variable are defined according to CIBSE Guide A [29], with increments of 1°C. The range extends from a setpoint of 18°C and a setback of 11°C to a warmer scenario with a setpoint of 23°C and a setback of 16°C.

### 4.4 Experimental tests

24 configurations of the sequential search are developed for experimental testing to assess its performance. These comprise four initial starting bounds, three variables grouping categories, and two algorithm runs (initial and iterative). This study investigates four different initial starting points initializing the search process with the variable values set at their low, middle, upper, and random bounds. Based on the variables detailed in section 4.3.2, the design options for these starting points are outlined in Table 4.

Table 4. Design options adopted for each initial starting point of the sequential search.

| | Variable | Low bound values | Middle bound values | Upper bound values | Random bound values |
|---|---|---|---|---|---|
| **Geometry** | Shape | Rectangle | L-shape | Free form | Free form |
| | Window-to-wall ratio (%) | 25 | 50 | 95 | 75 |
| | Orientation (˚) | Baseline (0) | 90 | 135 | 0 |
| **Fabric** | Thermal mass (-) | Lightweight | Heavyweight | Heavyweight | Lightweight |
| | Insulation thickness (cm) | (8.5) | + 50% (12.5) | 100% (17) | 25% (10.625) |
| | Window types (-) | Double clear + Air (2.8 $W/m^2K$) | Triple clear + Air (2 $W/m^2K$) | Triple LowE + Argon (1.23 $W/m^2K$) | Triple clear + Air (2 $W/m^2K$) |
| **HVAC** | Distribution type (-) | Radiant | Radiant | Forced air | Radiant |
| | Plant type (-) | Boiler | Heat pump | Heat pump | Boiler |
| | Supply water temperature (˚C) | 30 | 40 | 50 | 45 |
| **Controls** | Setpoint air temperature (˚C) | 18 | 20 | 23 | 21 |
| | Setback air temperature (˚C) | 11 | 13 | 16 | 15 |

Additionally, three different grouping categories are evaluated as experimental tests: ungrouped (without grouping the design variables), element-grouped (design variables grouped within the related design elements: geometry, fabric, HVAC systems and controls), and field-grouped (design variables grouped within the related field: architecture and engineering). Figure 5 illustrates these grouping categories with an example formulation of the design variables.

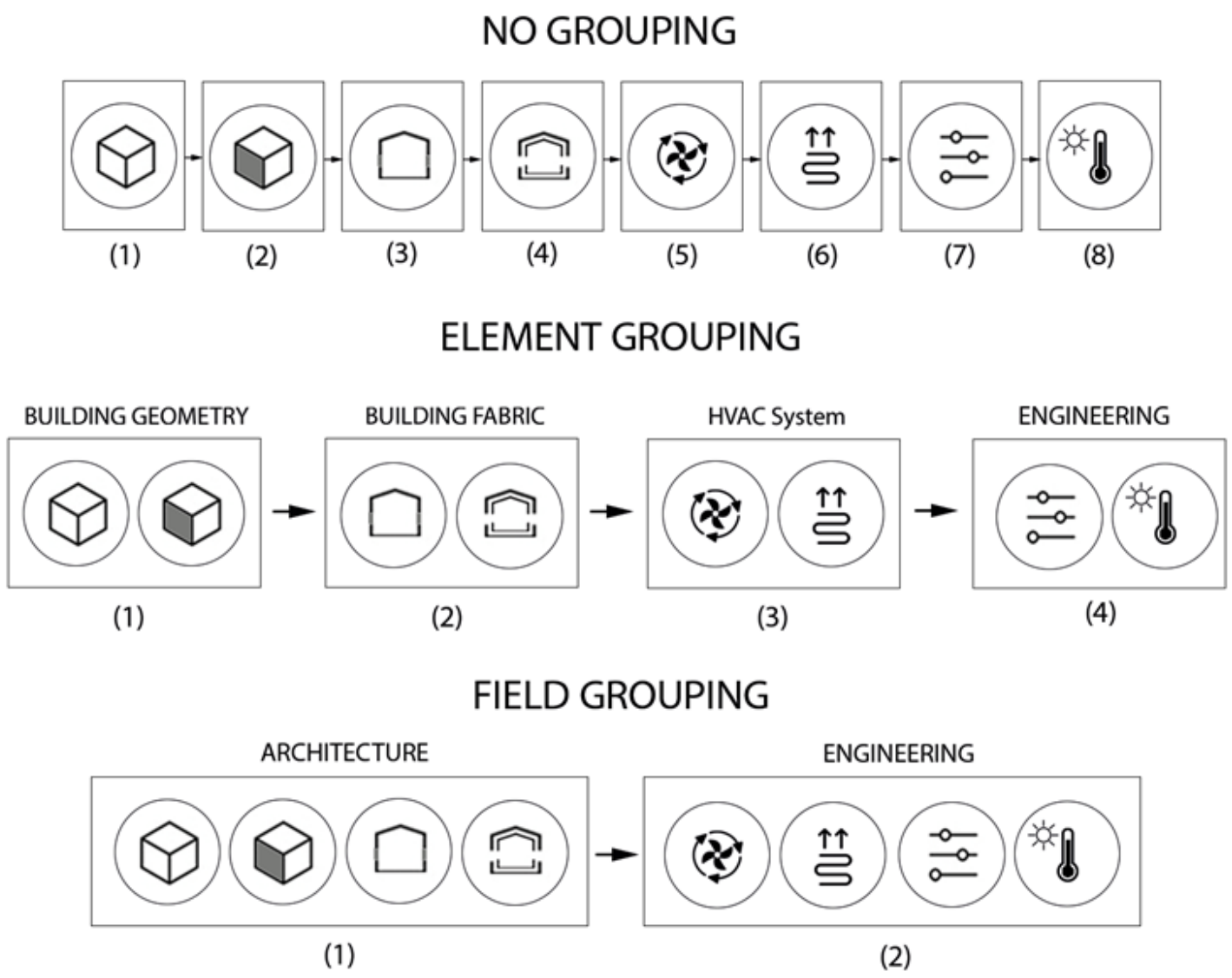

Figure 5. The ungrouped, element-grouped, and field-grouped categories.

The initial run terminates when the sequence of optimizations matches the number of stages dictated by the grouping category defined during initialization. The iterative run concludes once all Pareto optimal solutions from the initial run have been individually evaluated using the same sequential stages set during algorithm initialization. In this process, the Pareto optimal solutions identified in the initial run serve as starting points for the sequential search in the iterative run. Consequently, while the initial run begins with a single starting point, the iterative run commences with multiple starting points, corresponding to the number of optimal solutions found in the initial run.

## 4.5 Modelling and Simulation Implementation

The exhaustive dataset of solutions was generated using a two-step process (Figure 6). Grasshopper for Rhinoceros 3D parametrized building geometries and HVAC system variables, while jEPlus [35], a widely used IDF editor, managed variables directly configurable from IDF files (Figure x). This approach mitigates the latency of building performance simulations (BPS) in Grasshopper when handling large-scale simulations. Proprietary tools supported parametric modeling of building shapes and envelope geometries, with the Honeybee plug-in [31] translating geometries into thermal zones and assigning construction materials for EnergyPlus simulations via OpenStudio [33]. Ironbug enabled the design and sizing of HVAC systems and controls, while the TT Toolbox Iterator plug-in [34] generated OpenStudio Model (OSM) files for combinations of geometry and HVAC variables. These OSM files were converted into IDF blocks, which were stored as external files. A single IDF file was imported into jEPlus, where EP-Macro [36] incorporated stored IDF blocks along with remaining design variables to generate all variable combinations.

This process produced up to 4,147,200 simulations. Computational efficiency was achieved through concurrent processing, running 36 EnergyPlus simulations in parallel on an 18-core (36-thread) system. Each simulation covered an annual period using a Nottingham (UK) weather file, with six timesteps per hour for heat transfer and load calculations, as recommended by EnergyPlus. Simulation outputs, including energy demand and thermal discomfort results, were exported as CSV files. Post-processing in Python applied a Pareto ranking algorithm to identify optimal solutions from the full factorial search. These solutions were benchmarked against the optima from the sequential optimization algorithm and NSGA-II, both implemented in Python. The sequential approach implemented in this study is a unique, stand-alone code-based framework designed to optimize building performance through deterministic exhaustive (full factorial) searches within each stage.

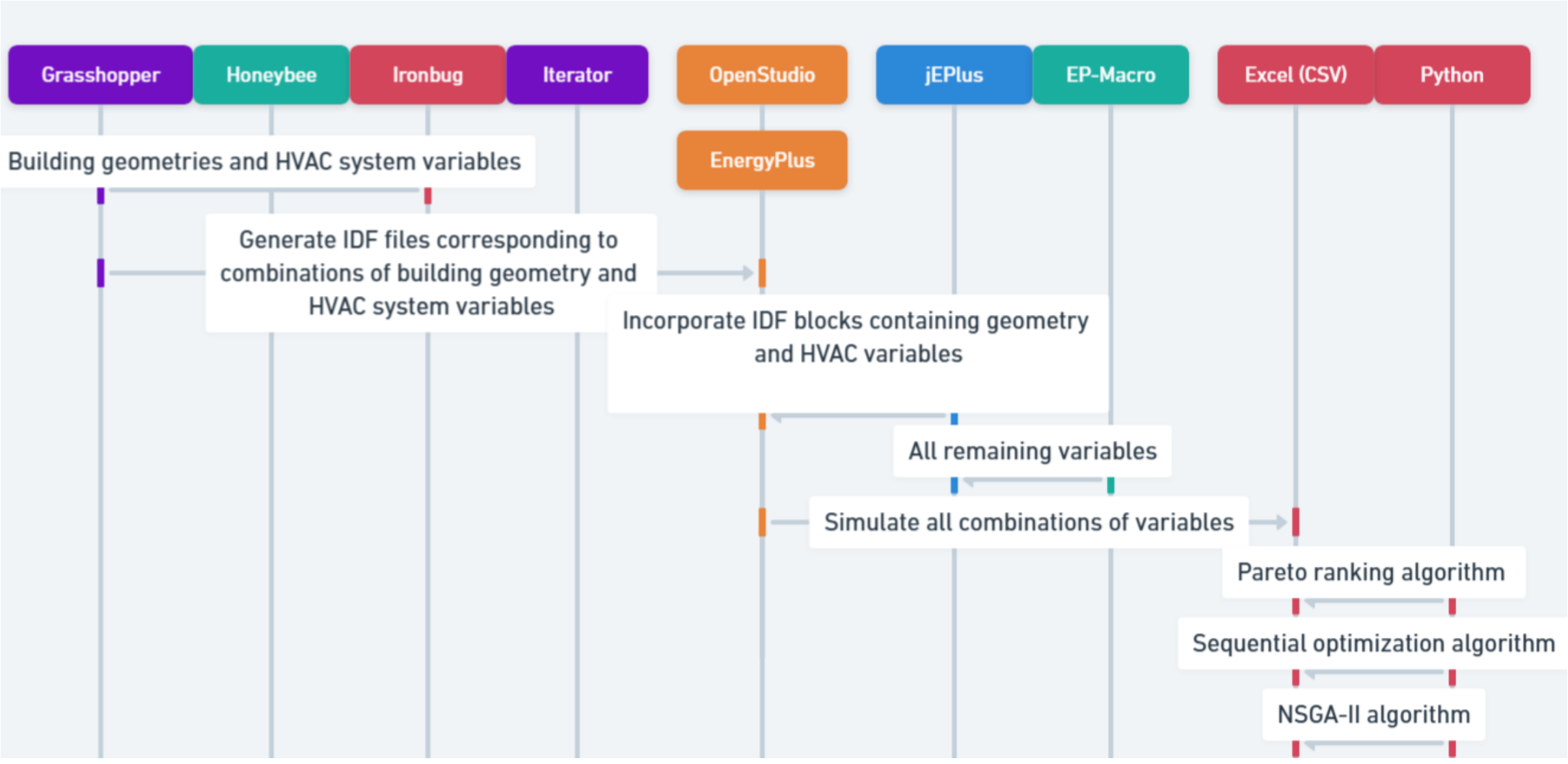

Figure 6. Tools and software employed for the generation of the solutions.

# 5. Results

The following sections report the outcomes of the sequential search. Section 5.1 describes the effectiveness and computational efficiency of the sequential optimal considering the large scale problem formulation. Section 5.2 describes the effect of problem size on the sequential optimization performance. Section 5.3 describes the effect of various weather and occupancy profiles on the sequential optimization performance. Section 5.4 lastly compares the performance of the sequential optimization to the commonly used NSGA-II algorithm.

## 5.1 Baseline Sequential Optimization Performance Comparison

This section describes the performance of the sequential optimization approach in detail for the large problem formulation.

### 5.1.1 Search effectiveness in finding global optimal solutions

Figure 7 details the number of search and global optimal solutions across experiments. Initializing the optimization from different regions of the solution space using various starting bounds does not impact the effectiveness of the ungrouped search. In fact, on average only one global optimum was identified across configurations of the sequential search, with the majority of the solutions differing from the global optima (search optima). However, a larger grouping of variables enhances the search effectiveness, as evidenced by an increased number of global optima found in the element and field grouped searches compared to the ungrouped experiment. While the mean effectiveness of the ungrouped search across starting bounds is only 1% relative to the full factorial search, it increases to 22.3% with the element grouped search and 59.8% with the field grouped configuration. Additionally, larger groupings of variables reduce the sensitivity to the starting bounds of the search, resulting in a smaller difference in the number of global optima found across configurations with different initial starting bounds. This is due to the mechanisms of the sequential search approach and the different ways each grouping strategy manages the search space and captures parameter interactions. Compared to the ungrouped search, grouped approaches expand the search space within each stage, allowing for a more comprehensive exploration of interdependencies among parameters, which results in a larger number of design combinations evaluated and consequently an increased number of optimal solutions generated and carried forward to the next optimization stage.

The iterative run increases the number of global optima obtained from the initial run for all variable grouping and initial starting bound experiments. This is the result of adopting the Pareto optimal solutions from the initial run as starting points, which allows for the exploration of regions in the solution space characterized by a higher number of well-performing solutions compared to the initial run. Across experiments with different starting bounds, the effectiveness of the ungrouped search increased from 1% to 46.4%, the element grouped search from 22.3% to 70.5%, and the field grouped search from 59.8% to 100%, in comparison to the full factorial search (100%). It is noticeable that starting the ungrouped or element grouped optimization from different regions of the solutions space greatly affects the number of global optimal solutions found (and the effectiveness of the search). However, the field grouped search identified the same number of global optimal solutions as the full factorial search (28), regardless of the initial starting point. Therefore, it can be considered 100% effective, irrespective of the initial starting point.

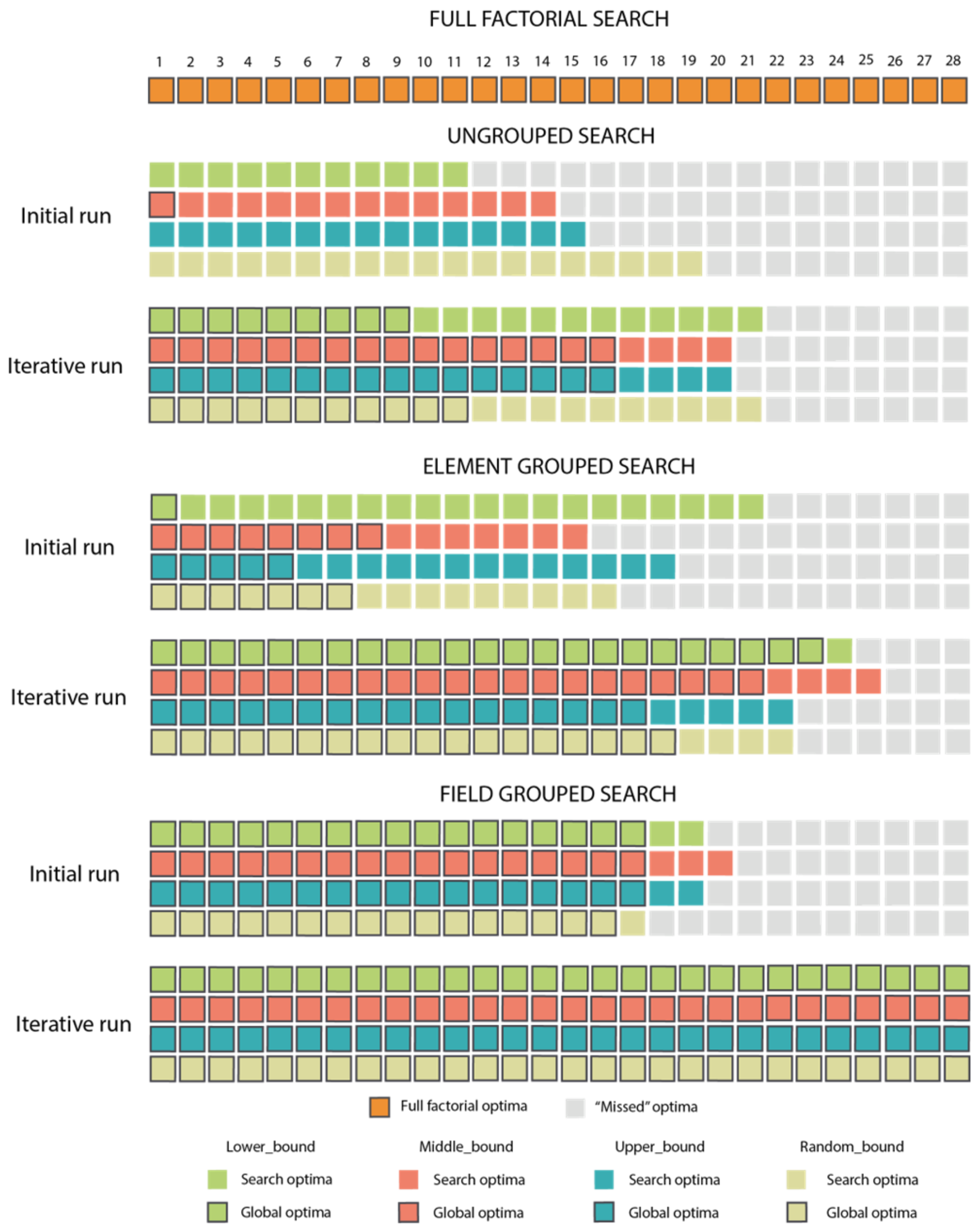

Figure 7. Search and global optima for the sequential search with different starting bounds and grouping categories concerning the initial and iterative runs, compared to the optimal solutions from the full factorial search.

### 5.1.2 Computational performance

In terms of computational performance, Figure 8 illustrates that the number of functions evaluations varies across sequential search experiments with different starting points, increasing as larger groupings of variables are explored. Initially, the ungrouped sequential search with various starting points evaluated between 119 and 243 function evaluations. Comparatively, the element grouped experiment evaluated between 538 and 755 function evaluations were evaluated during the element grouped experiment, increased to between 4,316 and 8,630 during the field grouped experiment. This corresponds to an average computational load of 0.02%, 0.06%, and 0.55%, respectively, relative to the full factorial approach (100%).

The increase in the number of global optima during the iterative run is achieved at the cost of higher computational loads for all starting points and grouping experiments. Following the initial run, the ungrouped sequential search with various starting points evaluated between 1,652 and 3,529 total combinations after the iterative run, compared to 7,892 – 10,602 for the element grouped experiment, and between 78,405 and 100,700 for the field grouped experiment. This results in an average computational load of 0.25%, 0.87%, and 8.79%, respectively, relative to the full factorial approach (100%). In summary, after the iterative run, the sequential search configurations offer computational savings between 99.8% and 89.5%, compared to a full factorial search. Specifically, the field grouped sequential search identified the same Pareto optimal solutions found during the full factorial search with a computational saving of 91.2%.

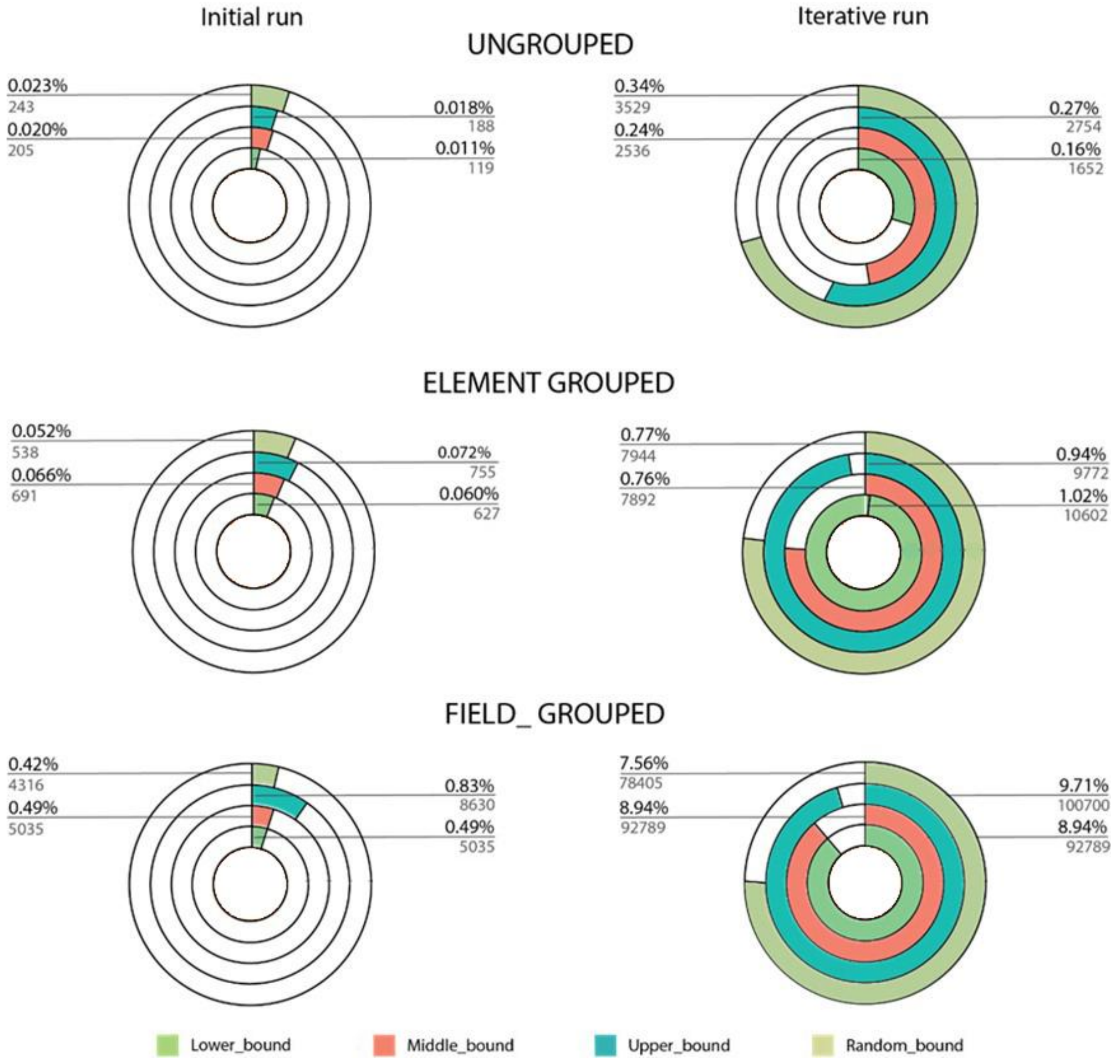

Figure 8. Computational performance of the sequential search configurations with different starting bounds and grouping categories for the initial run (left) and after the iterative run (right).

### 5.1.3 Relationship between search optima and global optima in the objective space

Despite both larger grouping categories and the iterative run increase the number of global optimal solutions obtained from the sequential search configurations, some search optima differ from the global optimal solutions derived from the full factorial search. Nevertheless, for all sequential search configurations, the comparison of trade-off values for energy demand and thermal comfort indicates performance values “close” to the global optimal set of solutions.

As portrayed in Figure 9, the percentage difference between the search and global optimal set diminishes with larger groupings of building variables (and an increase in number of function evaluations). Compared to the average percentage difference across ungrouped sequential search experiments, during the element grouped search, the percentage difference is reduced to 3.45% for energy demand and 4.18% for thermal comfort, resulting in a ~50% average reduction. Similarly, compared to the average percentage difference of the element grouped sequential search experiments, during the field grouped search, the percentage decreased by ~15%, to 2.6% for energy demand and 3.2% for thermal comfort. The observed percentage differences in performance metrics across grouping of variables are the results of search space inclusion and parameter interdependencies. The smaller search space of the ungrouped search and the limited exploration of interdependencies with other variables increases the likelihood of identifying optimal solutions that display larger deviation of performance metrics compared to the global optima. In contrast, the grouped approaches (element-grouped and field-grouped) incorporate broader sets of variables within each optimization stage, effectively expanding the search space at each step while accounting for synergies between design variables, resulting in optimal solutions that align closely with the performance metrics of the global optima.

A more significant reduction is observed during the iterative run where adopting the Pareto optimal solutions from the initial run as starting points enhances the exploration of regions in the solution space characterized by a higher number of well-performing solutions, leading to a ~60% average reduction in percentage difference for the ungrouped search (from 6.15% to 2.41% for energy demand and from 8.68% to 3.29% for thermal comfort). An additional ~20% reduction is achieved for the element grouped search with differences of 1.24% for energy demand and 3.29% for thermal comfort, and a further ~20% reduction during the field grouped search, where all the search optima match the energy and thermal comfort performances of the full factorial search set. Insights regarding the relationships between search and global optima in the variables space can be found in Appendix B.

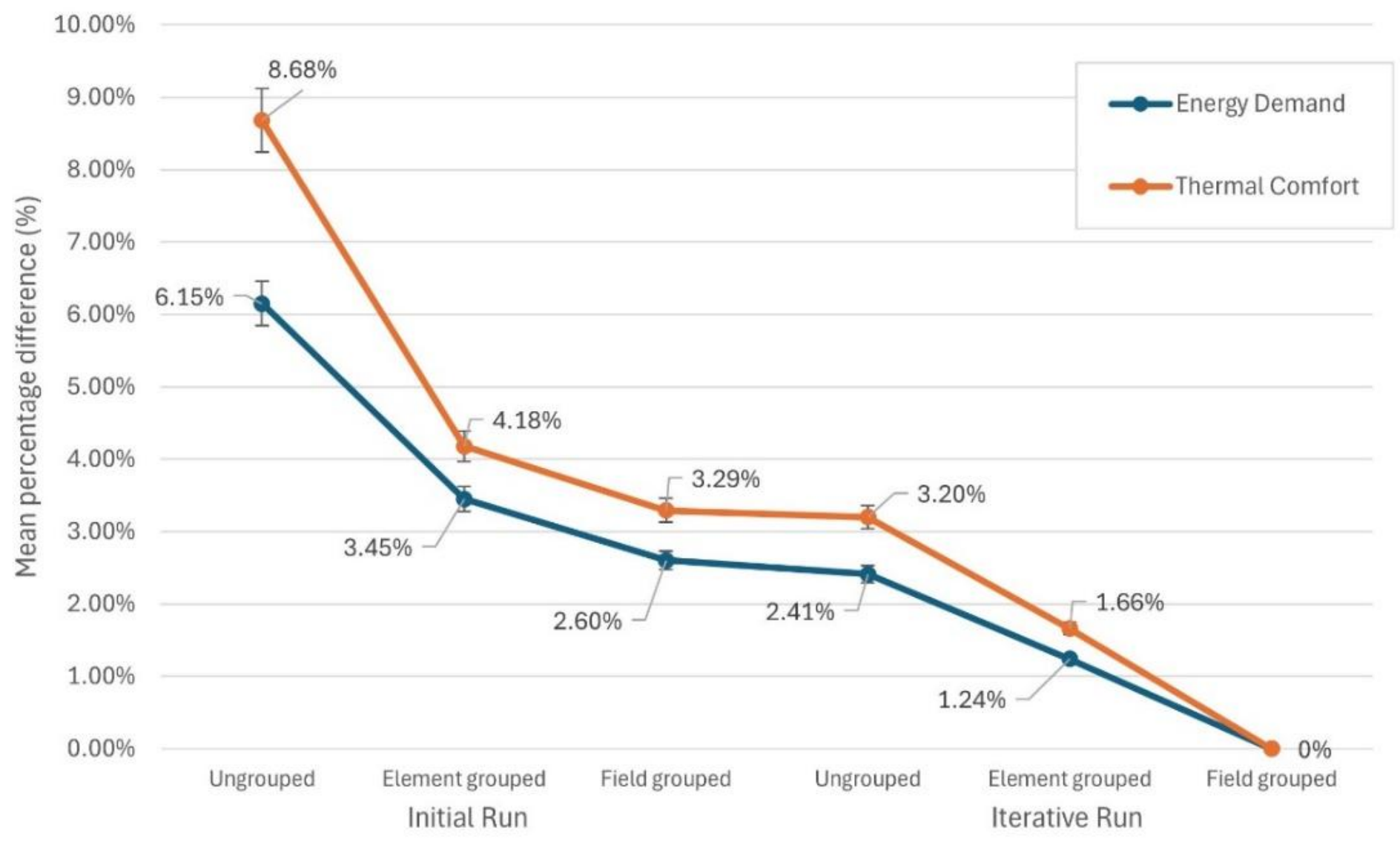

Figure 9. Mean percentage difference in energy demand and thermal comfort between the search optimal solutions obtained from different configurations of the sequential search and the full factorial.

## 5.2 Evaluating the effect of scale on the sequential optimization approach

This section describes the performance of the sequential algorithm across different problem scales specifically very small, small, medium, and large-scale problem.

### 5.2.1 Search effectiveness across different scales of problem formulations

As presented in Table 5, the field grouped sequential search during the iterative run consistently identified the global set of solutions identified by the full factorial search across all four scales of problem formulations, demonstrating 100% reliability. This indicates that the field grouped sequential search is insensitive to the initial starting bounds and maintains reliability across varying scales of problem formulations. Comparatively, the other configurations of the sequential search cannot be considered 100% reliable as their ability to identify the global optima is sensitive to the size of the problem formulation. Through a high-level comparison across scales, it is noticeable that the number of experimental tests that achieve the global optimal solutions decreases with larger problem formulations. Therefore, it can be concluded that the size of problem formulation affects the effectiveness of the sequential search to identify global optimal solutions across experimental tests. Specifically, for the very small problem formulation, all configurations of the sequential search (12 experiments in total) identified 4 global optima during iterative runs, matching the solutions found by the full factorial search. For the small-scale problem, the element and field grouped searches (8 experiments in total) identified 7 global optima. For medium and large-scale problems, only the field grouped searches achieved 21 and 28 global optima, respectively, during iterative runs, matching the solutions identified by the full factorial search.

Table 5. Number of global optima for the experimental tests of the sequential optimization across different scales of problem formulation, compared to the full factorial search.

| Scale | Runs | Full factorial | Ungrouped | | | | Element grouped | | | | Field grouped | | | |
|---|---|---|---|---|---|---|---|---|---|---|---|---|---|---|
| | | | **Low** | **Mid** | **Up** | **Rd** | **Low** | **Mid** | **Up** | **Rd** | **Low** | **Mid** | **Up** | **Rd** |
| *Very Small* | Initial | 4 | 3 | 2 | 2 | 1 | 4 | 3 | 3 | 2 | 4 | 4 | 4 | 4 |
| | Iterative | 4 | 4 | 4 | 4 | 4 | 4 | 4 | 4 | 4 | 4 | 4 | 4 | 4 |
| *Small* | Initial | 7 | 1 | 2 | 3 | 2 | 4 | 3 | 4 | 3 | 7 | 7 | 7 | 7 |
| | Iterative | 7 | 4 | 4 | 4 | 4 | 7 | 7 | 7 | 7 | 7 | 7 | 7 | 7 |
| *Medium* | Initial | 21 | 3 | 5 | 2 | 4 | 6 | 7 | 6 | 9 | 21 | 20 | 19 | 20 |
| | Iterative | 21 | 11 | 13 | 13 | 15 | 18 | 16 | 16 | 18 | 21 | 21 | 21 | 21 |
| *Large* | Initial | 28 | 0 | 1 | 0 | 0 | 1 | 8 | 5 | 7 | 17 | 17 | 17 | 16 |
| | Iterative | 28 | 9 | 16 | 16 | 11 | 23 | 21 | 17 | 18 | 28 | 28 | 28 | 28 |

### 5.2.2 Computational performance across different scales of problem formulations

Table 6 shows that an increase in number of function evaluations due to the larger scales of problem formulations results in a decrease in computational load with respect to the total solutions space evaluated by the full factorial search. This averages from 30.5% for the very small problem during the field grouped search, to 8.8% for the very large problem formulation. Consequently, computational savings relative to a full factorial search improve with increasing problem size. Specifically, the field grouped sequential search yields mean computational savings of approximately 70% for the very small problem, which increase to 78.6% for the small problem, 86.5% for the medium problem, and 91.2% for the large problem formulation.

Table 6. Computational load for the experimental tests of the sequential optimization across different scales of problem formulations, compared to the full factorial search.

| Scale | Full factorial | Ungrouped (%) | | | | Element grouped (%) | | | | Field grouped (%) | | | |
|---|---|---|---|---|---|---|---|---|---|---|---|---|---|
| | | **Low** | **Mid** | **Up** | **Rd** | **Low** | **Mid** | **Up** | **Rd** | **Low** | **Mid** | **Up** | **Rd** |
| V.Small | 100 | 9.9 | 5.9 | 5.9 | 6.5 | 11.3 | 9.9 | 10.2 | 10.2 | 30.5 | 30.5 | 30.5 | 30.5 |
| Small | 100 | 1.6 | 1.7 | 1.7 | 1.7 | 5.8 | 5.8 | 6.7 | 5.9 | 21.4 | 21.4 | 21.4 | 21.4 |
| Medium | 100 | 0.5 | 0.8 | 1.4 | 1.3 | 1.6 | 3 | 3.4 | 3.4 | 12.3 | 13.3 | 13.7 | 14.4 |
| Large | 100 | 0.1 | 0.2 | 0.3 | 0.4 | 1 | 0.8 | 0.9 | 0.8 | 8.9 | 8.9 | 9.7 | 7.6 |

## 5.3 Search reliability across variations of problem formulations

To understand if the performance of the sequential search algorithm varied across variations of problem formulations, the sequential approach was evaluated for the large-scale problem considering two weather scenarios and occupancy conditions. As detailed in Figure 10, the field grouped sequential search during the iterative run consistently exhibited a 100% effectiveness in identifying the global set of solutions obtained by the full factorial search across variations of problem formulations, demonstrating 100% reliability. This indicates that the field grouped sequential search maintains reliability across varying problem formulations. Comparatively, the other configurations of the sequential search cannot be considered 100% reliable as their effectiveness is sensitive to the characteristics of the problem formulation. However, the deviation in reliability varies depending on the configurations of the sequential search and increases with larger grouping of variables as proxies for a larger search space explored and consequently a larger number of global optima identified. Compared to the low deviation displayed by the ungrouped search across variations of the same underlined problem formulation (4.3%), the deviation across problem formulations for the element and field grouped search increased to 5.2% and 15.9%, respectively. The iterative run increased the deviation in effectiveness across variations of problem formulations of the ungrouped and element grouped search increased to 12.3% and 20.5%, respectively.

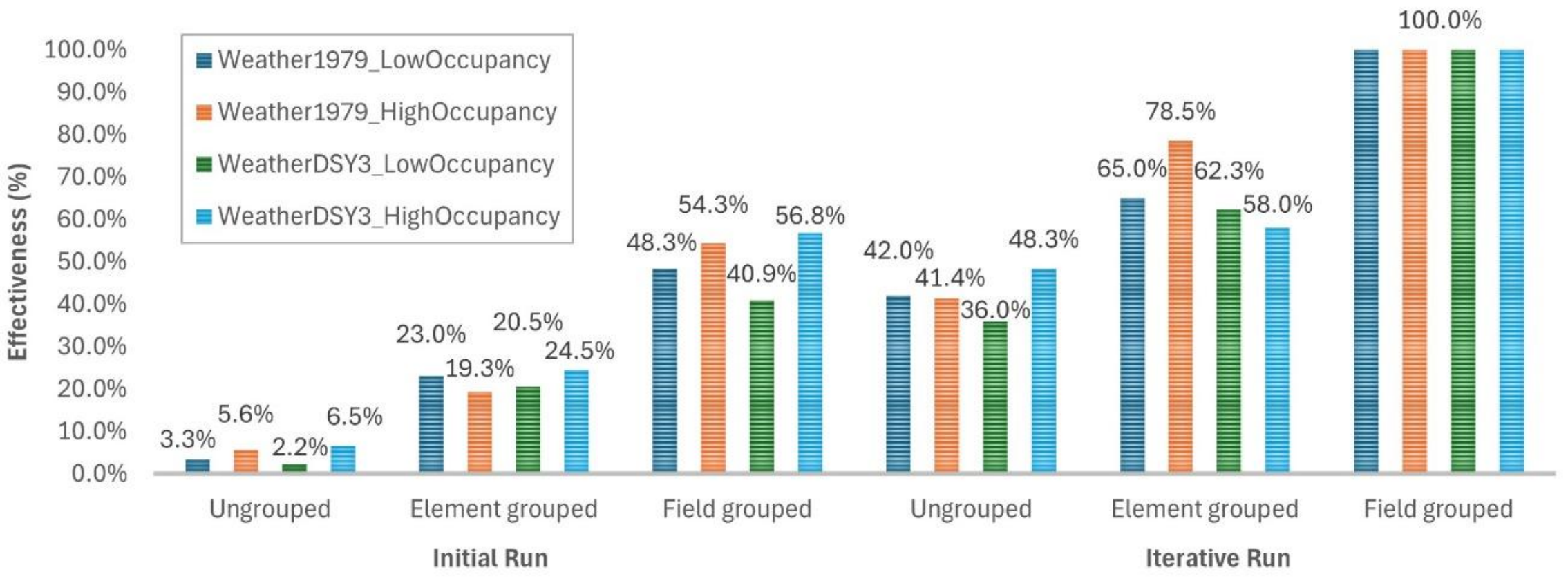

Figure 10. Mean search effectiveness in finding global optimal solutions across variations of the same problem formulation derived by uncertain boundary conditions, compared to the full factorial search.

## 5.4 Evaluating the efficiency of the sequential optimization approach as compared to NSGA II

The performance of the sequential search was compared to the NSGA II algorithm by fixing the number of function evaluations for NSGA II to those achieved by the sequential approach. The sections below compare the effectiveness and performance of both approaches.

### 5.4.1 Search efficiency in finding global optimal solutions

Generally, both the experiments of the sequential search and the NSGA-II display an increased effectiveness in finding global optimal solutions with an increased number of function evaluations (Figure 11). Up to 243 function evaluations, the ungrouped sequential search and the NSGA-II exhibited the same effectiveness in finding global optimal solutions. In fact, on average, only one global optimum was found among the configurations of the ungrouped sequential search with different starting points and across repeated runs of the GA (1% effectiveness). However, between 755 and 100700 function evaluations, all the configurations of the sequential search outperformed the NSGA-II. With 755 function evaluations, the element grouped search (22.3%) is three times as effective as the GA (7.2%) in identifying global optimal solutions. By significantly increasing the number of function evaluations to 8630, the field grouped search (59.8%) exhibits a percentage of effectiveness twice the one obtained from the NSGA-II algorithm (32.2%). Although a larger number of function evaluations enhances the effectiveness of the NSGA-II as evidenced by a closer performance to the experiments of the sequential search, the field grouped search during the iterative run identified the same number of global optimal solutions as the full factorial search (100% effective), regardless of the initial starting point, with 100700 function evaluations. With the same amount of function evaluations, the NSGA-II obtained 73.5% of the global optima found from the full factorial search. Therefore, for the same amount of function evaluations, the sequential search appears to be more efficient that the GA, irrespective of the starting points. Finally, regardless of the amount of function evaluations, compared to the stochastic (probabilistic) nature of the NSGA-II which results in in different solutions being found across repeated runs of the algorithm, the sequential search offered consistency across repeated runs of the search since it is based on full factorial sequential optimization steps and can be therefore considered deterministic.

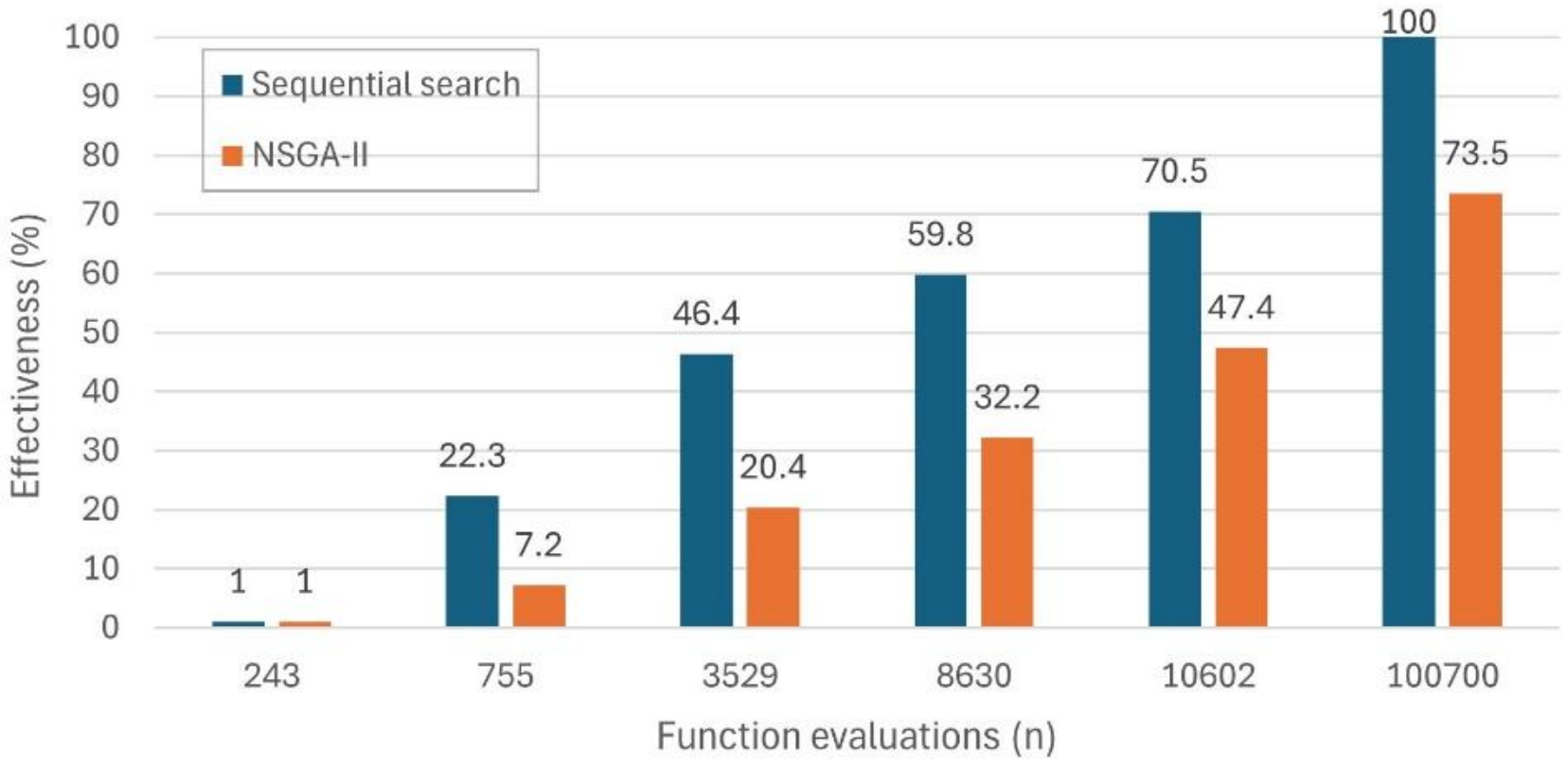

Figure 11. Compared search effectiveness between the sequential optimization and the NSGA-II with an increasing number of function evaluations.

### 5.4.2 Relationship between search optima and global optima in the objective space

Generally, all the configurations of the sequential search outperformed the NSGA-II in terms of mean percentage difference in energy and thermal comfort with respect to the full factorial solutions (Figure 12). A longer computational time increases the ability of both the sequential search and the NSGA-II to match the energy and thermal comfort performance obtained from a full factorial search. However, the improvement of the sequential search with an increased amount of function evaluations is much more significant compared to the NSGA-II. In fact, the mean percentage difference in energy demand and thermal comfort with respect to the full factorial search for the NSGA-II decreases by about 50% from 8.68% to 4.17% and from 11.32% to 6.35%, respectively. Comparatively, the sequential search after 100700 function evaluations matches the energy and thermal comfort performance obtained by the full factorial search from an initial percentage difference of 6.15% and 8.46%, respectively. This results in a larger gap of performance between the sequential search and the NSGA-II with an increased computational time, with the latter displaying differences in performance with a larger number of function evaluations. In fact, while with 243 function evaluations, the performance gap between the sequential search and the NSGA-II is between 2%-3%, it increases to 4%-6% with 100,700 function evaluations.

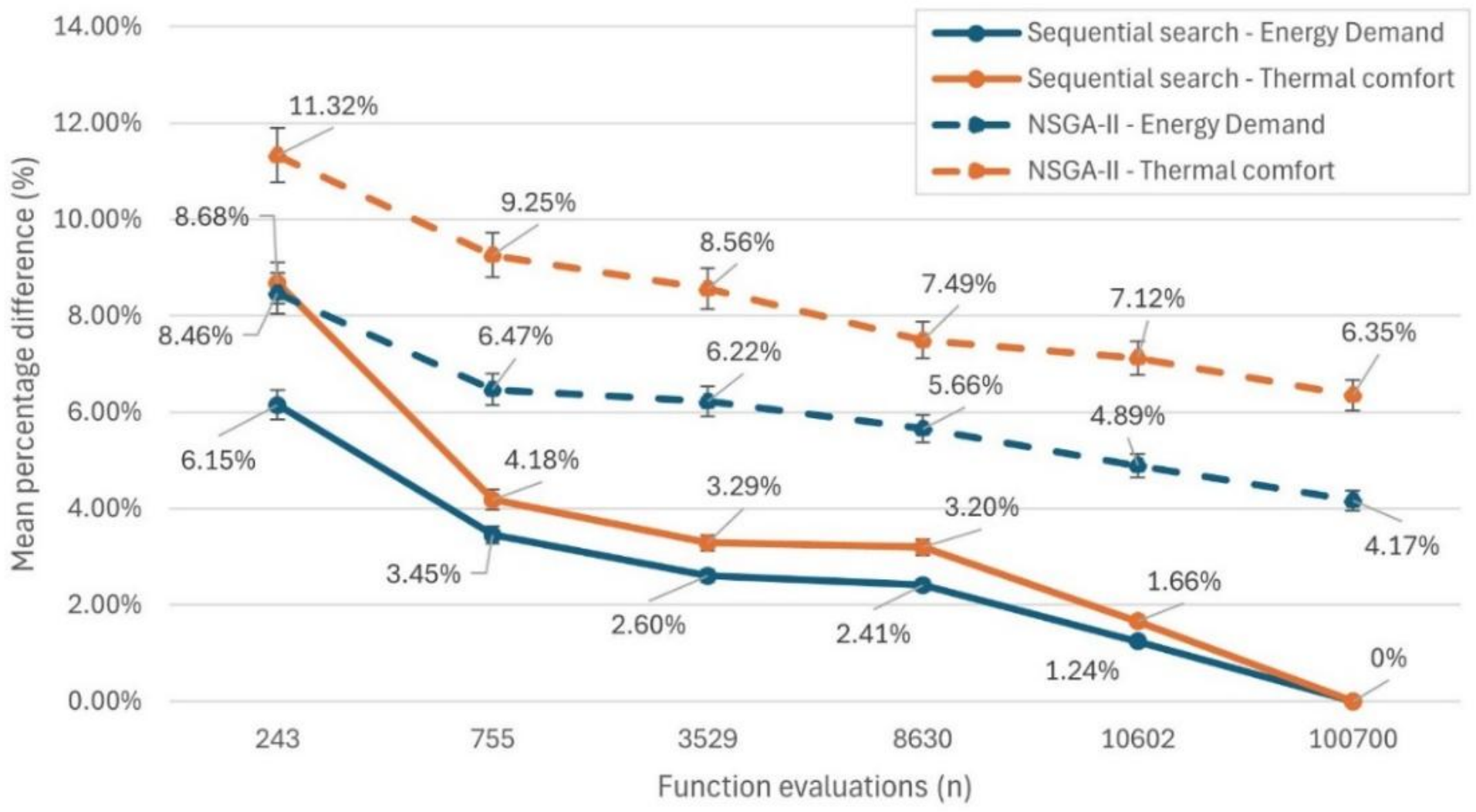

Figure 12. Compared mean percentage difference in energy demand and thermal comfort between the search optimal solutions obtained from the sequential search and the NSGA-II and the full factorial.

## 6. Discussion

The evaluation of the sequential optimization approach presented in this study represents a significant advancement in addressing the key challenges of performance-based building design. The approach 1) offers a reliable holistic optimization of building design elements while capturing their interdependencies, 2) reduces the computational effort needed for building performance evaluation, 3) simplifies the integration of optimization into design workflows empowering building stakeholders to adopt performance-based strategies in their designs. Among the experimental configurations of the sequential search, the field-grouped sequential search during the iterative run, a two-stage process termed field-grouped involving sequential optimizations of (building geometry + fabric) and (HVAC system + controls), identified the same Pareto optimal solutions as a full factorial search, across all four scales and four variations of problem formulations. Additionally, the consistency in results is obtained while achieving computational savings of 91.2% in function evaluations compared to a full factorial search. The differences in effectiveness, reliability and efficiency between the ungrouped and grouped approaches highlight the importance of search space expansion and variables interdependency in the sequential search approach configurations. The grouped configurations enhance search performance by allowing simultaneous optimization of multiple parameters within each stage, enabling the approach to leverage interactions between multiple design elements within an increased search space. By contrast, the limited search space of the ungrouped approach and its individual parameter focus limit the ability to explore complex interactions, leading to higher variability in solutions and reduced performance.

Finally, the performance comparison against the widely popular NSGA-II algorithm demonstrated the higher efficiency of the field-grouped sequential search in identifying optimal solutions. In fact, with the same amount of function evaluations (100,700), the NSGA-II obtained only 73.5% of the global optima found from the full factorial search.

Furthermore, the results indicate that the sequential search can be initiated from any region of the search space (or any selected initial solution) without compromising performance. Although the various configurations of the sequential search demonstrated lower effectiveness in identifying global optimal solutions compared to a full factorial, they remain viable for adoption by the building stakeholders depending on the scope of the optimization or at various stages of the building design process. This is due to (1) their performance values being "close" to the global set of optimal solutions, coupled with significantly reduced computational load. The smallest mean differences between the search optima and the global optimal solution set were 1.24% for energy demand and 1.66% for thermal comfort, while the largest differences were 6.15 % for energy demand and 8.68% for thermal comfort achieved with an average computational load of 0.35% of the full factorial search. (2) Their capability to identify the same design options obtained from the full factorial search for all design parameters with the highest impact on both energy demand and thermal comfort, as these design options exhibit high sensitivity to the optimization objectives.

Although this study utilizes a specific building type and a defined set of variables and objectives to investigate the performance of the sequential optimization process, the approach is inherently adaptable. The approach is not dependent on the unique characteristics of any single building configuration; rather, it leverages a generalizable sequence of optimization stages—covering geometry, fabric, HVAC, and controls—that can be applied to diverse building types, climate conditions, and performance objectives. This versatility makes the method applicable across a range of architectural and engineering contexts, supporting customized, performance-driven design solutions.

## 6.1 Practical use of the proposed sequential optimization approach

The results indicate that integrating the sequential design optimization approach within the building design process can proactively facilitate the identification of well performing designs. However, building stakeholders must operate in a collaborative environment where design alternatives are proposed through interdisciplinary knowledge. Furthermore, stakeholders are required to adopt a holistic-thinking approach, enhancing opportunities for synergies between building design elements and considering the resulting complex interactions. Finally, at the conclusion of the process, the design team can jointly select one or more satisfactory design solutions from the final set of Pareto optimal solutions, which encompass both architectural (building geometry and fabric) and technical (HVAC system and controls) features and refine it during the forthcoming detailed stages.

The results of this paper provide detailed insights on the landscape of configurations of the sequential search. However, building stakeholders are required to evaluate the trade-offs between the benefits and limitations of each configuration based on the presented results. While the field-grouped sequential search during the iterative run matched the performance results of a full factorial, all configurations of the sequential search exhibited varying magnitudes of performance. Therefore, building stakeholders can adopt any of these configurations depending on the scope of the optimization. For instance, the design team might prioritize a lower computational load over a higher effectiveness in finding global optimal solutions. Conversely, building stakeholders might opt to perform both the initial and iterative runs of the search to achieve high effectiveness in identifying global optimal solutions, thereby justifying the higher computational load. Alternatively, different configurations of the sequential search can be adopted at various stages of the building design process. The ungrouped sequential search, despite its higher sensitivity to starting points and lower magnitude of performance, could provide a rapid, lower-complexity tool to facilitate the early identification of key design parameters influencing building performance metrics. This setup can support concept-stage exploration by allowing building stakeholders to identify and prioritize the parameters with the greatest influence on performance objectives, guiding more detailed exploration in subsequent stages. In contrast, the element-grouped search, which has a higher computational load but greater performance, could be employed during the exchange of information between the architectural and engineering teams. This would facilitate the joint selection of variables related to building design elements and their perturbation values for further optimization assessments. Finally, the field-grouped sequential search could be utilized during the final stages of design process, prior to the technical design, to identify the global optima. These solutions would then be assessed during the decision-making process, culminating in the selection of the final design.

## 7. Conclusion

This study addresses key gaps in existing literature by proposing a novel sequential optimization approach that harnesses the synergies and interdependencies between building design elements, while reducing both the complexity of the algorithmic process and the computational effort required for performance evaluation. A benchmarking framework is established to evaluate its performance against a full factorial search, focusing on the effectiveness in identifying global optimal solutions, computational efficiency, and solution optimality. In addition, the reliability of the sequential search is assessed across four distinct scales and four variations of a problem formulations ranging from 874 to 4,147,200 design options for an office building. Furthermore, the efficiency of the sequential search is benchmarked against the widely recognized Non-Dominated Sorting Genetic Algorithm (NSGA-II). To validate its performance, 24 distinct configurations of the sequential search algorithm were developed as part of the experimental tests.

The proposed two-stage, field-grouped sequential search, applied to building geometry, fabric, HVAC systems, and controls, consistently identified the same Pareto-optimal solutions as a full factorial search across different problem scales, demonstrating 100% reliability and effectiveness. This approach achieved a 91.2% reduction in function evaluations compared to a full factorial search, highlighting its computational efficiency. Furthermore, in comparison with the widely used NSGA-II algorithm, the sequential approach outperformed in identifying global optima. Although, the findings underscore the superior efficiency, reliability, and effectiveness of the field-grouped sequential search, building stakeholders can leverage various configurations of the search to explore design alternatives and optimize building geometry, fabric, HVAC system and controls during various stages of the building design process, assessing the trade-off between results accuracy and computational overhead.

This study provides a clear roadmap for replicating the sequential optimization approach, combining detailed methodological explanations, pseudocode, and adaptable design stages that allow it to be applied in various contexts. By supporting adjustments for specific building types, climates, and design variables and objectives, the sequential optimization approach offers a practical and scalable solution for building stakeholders seeking data-driven, performance-oriented outcomes.

## CRediT authorship contribution statement

**Riccardo Talami:** Conceptualization, Methodology, Software, Validation, Formal analysis, Visualization, Data Curation, Writing – Original Drafts, Writing – Review & Editing, Visualization. **Jonathan Wright:** Conceptualization, Methodology, Resources, Writing – Original Draft, Writing – Review & Editing, Supervision, Project administration, Funding acquisition. **Bianca Howard:** Conceptualization, Methodology, Resources, Writing – Original Draft, Writing – Review & Editing, Supervision, Project administration, Funding acquisition.

## Declaration of competing interest

The authors declare that they have no known competing financial interests or personal relationships that could have appeared to influence the work reported in this paper.

## Acknowledgements

This research is supported through a scholarship funded by the UK Engineering and Physical Sciences Research Council (EPSRC) and a philanthropic donation from J. Atack and J. Taylor.

## Appendix A

Table A.1. Fixed input parameter values in EnergyPlus for office spaces.

| **Object** | **Parameter** | **Value** |
|---|---|---|
| Occupant density | People per zone floor area (persons/m$^2$)<br>Sensible heat fraction | 0.1<br>0.5935 |
| Light density | Watts per zone floor area (W/m$^2$). Dimming daylighting control with a setpoint of 300 lux at a reference point located in the middle of the space. | 9 |
| Electric equipment load | Watts per zone floor area (W/m$^2$)<br>Fraction latent | 11<br>0.75 |
| Infiltration rate | Air changes per hours (ac/h) | 0.5 |
| Outdoor air flow | Outdoor airflow per person (m$^3$/s person) | 0.01 |
| Activity | Activity level (W/m$^2$) | 65 |

Table A.2. Schedule of the input parameters in EnergyPlus for office spaces.

| **Activity level** | **People** | **Lights** | **Electric equipment** |
|---|---|---|---|
| Through: 12/31<br>For: all days<br>Until: 24:00<br>117 (typing), 1.1 met | Through: 12/31<br>For: Weekdays<br>Until: 07:00<br>0<br>Until: 08:00<br>0.2<br>Until: 09:00<br>0.6<br>Until: 12:00<br>1<br>Until: 14:00<br>0.8<br>Until: 17:00<br>1<br>Until: 18:00<br>0.6<br>Until: 19:00<br>0.2<br>Until: 24:00<br>0<br>For: Other days<br>Until: 24:00<br>0 | Through: 12/31<br>For: Weekdays<br>Until: 07:00<br>0<br>Until: 19:00<br>1<br>Until: 24:00<br>0<br>For: Other days<br>Until: 24:00<br>0 | Through: 12/31<br>For: Weekdays<br>Until: 07:00<br>0.05<br>Until: 19:00<br>1<br>Until: 24:00<br>0.05<br>For: Other days<br>Until: 24:00<br>0.05 |

Table A.3. Building design variables and option values.

| | Variable | Option | Vsmall | Small | Medium | Large |
|---|---|---|---|---|---|---|
| **Geometry [28]** | Shape | Rectangle | ● | ● | ● | ● |
| | | L-shape | | ● | ● | ● |
| | | Free form | ● | ● | ● | ● |
| | Window-to-wall ratio (%) | 25 | ● | ● | ● | ● |
| | | 50 | ● | ● | ● | ● |
| | | 75 | | | ● | ● |
| | | 95 | ● | ● | ● | ● |
| | Orientation (˚) | 0 | | ● | ● | ● |
| | | 45 | | ● | ● | ● |
| | | 90 | | ● | ● | ● |
| | | 135 | | | | ● |
| **Fabric [18,29]** | Thermal mass (-) | Lightweight | ● | ● | ● | ● |
| | | Heavyweight | ● | ● | ● | ● |
| | Insulation thickness (cm) | 8.5 | | ● | ● | ● |
| | | +25% (10.625) | | | ● | ● |
| | | + 50% (12.5) | | ● | ● | ● |
| | | +75% (15.625) | | | ● | ● |
| | | +100% (17) | | ● | ● | ● |
| | Window types (-) | Double clear + Air (2.8 W/m$^2$K) | ● | ● | ● | ● |
| | | Triple clear + Air (2 W/m$^2$K) | ● | ● | ● | ● |
| | | Triple LowE + Argon (1.23 W/m$^2$K) | | ● | ● | ● |
| **HVAC [19, 26, 27]** | Distribution type (-) | Radiant | ● | ● | ● | ● |
| | | Forced air | ● | ● | ● | ● |
| | Plant type (-) | Boiler | ● | ● | ● | ● |
| | | Heat pump | ● | ● | ● | ● |
| | Supply water temperature (˚C) | 30 | | ● | ● | ● |
| | | 35 | | | ● | ● |
| | | 40 | | ● | ● | ● |
| | | 45 | | | ● | ● |
| | | 50 | | ● | ● | ● |
| **Controls [19, 20]** | Setpoint air temperature (˚C) | 18 | | ● | | ● |
| | | 19 | ● | | ● | ● |
| | | 20 | | ● | | ● |
| | | 21 | ● | | ● | ● |
| | | 22 | | ● | | ● |
| | | 23 | ● | | ● | ● |
| | Setback air temperature (˚C) | 11 | | ● | ● | ● |
| | | 12 | ● | | | ● |
| | | 13 | | ● | ● | ● |
| | | 14 | ● | | | ● |
| | | 15 | | ● | ● | ● |
| | | 16 | ● | | | ● |

Table A.4. Material layers and thermophysical properties of external wall constructions.

| Layer | | Heavyweight | Lightweight |
|---|---|---|---|
| **1 (out)** | *Material*<br>*Thickness (m)*<br>*Thermal conductivity (W/mK)*<br>*Density (kg/m³)*<br>*Specific heat capacity (J/kgK)* | Plaster<br>0.013<br>0.220<br>800<br>840 | Shingles<br>0.010<br>0.120<br>510<br>1260 |
| **2** | *Material*<br>*Thickness (m)*<br>*Thermal conductivity (W/mK)*<br>*Density (kg/m³)*<br>*Specific heat capacity (J/kgK)* | Expanded polystyrene board<br>0.085<br>0.033<br>15<br>1450 | Battens (Air gap) |
| **3** | *Material*<br>*Thickness (m)*<br>*Thermal conductivity (W/mK)*<br>*Density (kg/m³)*<br>*Specific heat capacity (J/kgK)* | Concrete<br>0.180<br>1.300<br>2000<br>840 | Chipboard<br>0.012<br>0.140<br>600<br>1700 |
| **4** | *Material*<br>*Thickness (m)*<br>*Thermal conductivity (W/mK)*<br>*Density (kg/m³)*<br>*Specific heat capacity (J/kgK)* | | Wool insulation<br>0.075<br>0.039<br>25<br>1800 |
| **5** | *Material*<br>*Thickness (m)*<br>*Thermal conductivity (W/mK)*<br>*Density (kg/m³)*<br>*Specific heat capacity (J/kgK)* | | Chipboard<br>0.012<br>0.140<br>600<br>1700 |
| **6** | | | Vapour barrier |
| **7** | | | Battens (Air gap) |
| **8 (in)** | *Material*<br>*Thickness (m)*<br>*Thermal conductivity (W/mK)*<br>*Density (kg/m³)*<br>*Specific heat capacity (J/kgK)* | | Plaster board<br>0.013<br>0.210<br>700<br>1000 |

Table A.5. Material layers and thermophysical properties of roof constructions.

| Layer | | Heavyweight | Lightweight |
|---|---|---|---|
| **1 (out)** | *Material*<br>*Thickness (m)*<br>*Thermal conductivity (W/mK)*<br>*Density (kg/m³)*<br>*Specific heat capacity (J/kgK)* | Shingles<br>0.010<br>0.120<br>510<br>1260 | Shingles<br>0.010<br>0.120<br>510<br>1260 |
| **2** | *Material* | Battens (Air gap) | Battens (Air gap) |
| **3** | *Material* | Breather membrane | Breather membrane |
| **4** | *Material*<br>*Thickness (m)*<br>*Thermal conductivity (W/mK)*<br>*Density (kg/m³)*<br>*Specific heat capacity (J/kgK)* | Fibreboard<br>0.019<br>0.082<br>350<br>1300 | Chipboard<br>0.012<br>0.140<br>600<br>1700 |
| **5** | *Material*<br>*Thickness (m)*<br>*Thermal conductivity (W/mK)*<br>*Density (kg/m³)*<br>*Specific heat capacity (J/kgK)* | Expanded polystyrene board<br>0.110<br>0.033<br>15<br>1450 | Wool insulation<br>0.125<br>0.039<br>25<br>1800 |
| | *Material* | Concrete | Chipboard |
| **6** | *Thickness (m)*<br>*Thermal conductivity (W/mK)*<br>*Density (kg/m$^3$)*<br>*Specific heat capacity (J/kgK)* | 0.200<br>1.300<br>2000<br>840 | 0.012<br>0.140<br>600<br>1700 |
| **7** | *Material* | | Vapour barrier |
| **8** | *Material* | | Battens (Air gap) |
| **9 (in)** | *Material*<br>*Thickness (m)*<br>*Thermal conductivity (W/mK)*<br>*Density (kg/m$^3$)*<br>*Specific heat capacity (J/kgK)* | | Plaster board<br>0.013<br>0.210<br>700<br>1000 |

Table A.6. Material layers and thermophysical properties of internal wall and foundation slab.

| | | | |
|---|---|---|---|
| **1 (out)** | *Material*<br>*Thickness (m)*<br>*Thermal conductivity (W/mK)*<br>*Density (kg/m$^3$)*<br>*Specific heat capacity (J/kgK)* | Plaster board<br>0.013<br>0.210<br>700<br>1000 | Granular fill<br>0.150<br>0.360<br>1840<br>840 |
| **2** | *Material*<br>*Thickness (m)*<br>*Thermal conductivity (W/mK)*<br>*Density (kg/m$^3$)*<br>*Specific heat capacity (J/kgK)* | Metal studs<br><br>0.050<br><br> | Cement screed<br>0.150<br>1.400<br>2100<br>650 |
| **3** | *Material*<br>*Thickness (m)*<br>*Thermal conductivity (W/mK)*<br>*Density (kg/m$^3$)*<br>*Specific heat capacity (J/kgK)* | Plaster board<br>0.013<br>0.210<br>700<br>1000 | Vapour barrier |
| **4** | *Material*<br>*Thickness (m)*<br>*Thermal conductivity (W/mK)*<br>*Density (kg/m$^3$)*<br>*Specific heat capacity (J/kgK)* | | Expanded polystyrene<br>0.200<br>0.033<br>15<br>1450 |
| **5** | *Material* | | Vapour barrier |
| **6 (in)** | *Material*<br>*Thickness (m)*<br>*Thermal conductivity (W/mK)*<br>*Density (kg/m$^3$)*<br>*Specific heat capacity (J/kgK)* | | Concrete<br>0.200<br>1.300<br>2000<br>840 |

Table A.7. Material layers and thermophysical properties of window types.

| **Layer** | | Double pane + Air | Triple Pane + Air | Triple Pane LowE + Argon |
|---|---|---|---|---|
| **1 (out)** | *Material*<br>*Thickness (m)*<br>*Thermal conductivity (W/mK)*<br>*Solar transmittance*<br>*Visible transmittance* | Clear pane<br>0.003<br>0.9<br>0.630<br>0.850 | Clear pane<br>0.003<br>0.9<br>0.630<br>0.850 | Clear pane<br>0.003<br>0.9<br>0.630<br>0.850 |
| **2** | *Material*<br>*Thickness (m)* | Air<br>0.008 | Air<br>0.008 | Argon<br>0.013 |
| **3** | *Material*<br>*Thickness (m)*<br>*Thermal conductivity (W/mK)*<br>*Solar transmittance*<br>*Visible transmittance* | Clear pane<br>0.003<br>0.9<br>0.630<br>0.850 | Clear pane<br>0.003<br>0.9<br>0.630<br>0.850 | Clear pane<br>0.003<br>0.9<br>0.630<br>0.850 |
| **4** | *Material*<br>*Thickness (m)* | | Air<br>0.008 | Argon<br>0.013 |
| **5** | *Material*<br>*Thickness (m)*<br>*Thermal conductivity (W/mK)*<br>*Solar transmittance*<br>*Visible transmittance* | | Clear pane<br>0.003<br>0.9<br>0.630<br>0.850 | Low emissive coating |
| **6 (in)** | *Material*<br>*Thickness (m)*<br>*Thermal conductivity (W/mK)*<br>*Solar transmittance*<br>*Visible transmittance* | | | Clear pane<br>0.003<br>0.9<br>0.630<br>0.850 |

# Appendix B

## Relationship between search and global optima in the variables space

Figure B.1 depicts the trends of design variables identified as optimal across sequential search experiments. The thickness of each line represents the frequency with which variables values appear in the optimal solutions across the initial starting points (low, middle, upper, and random). Thicker lines indicate higher consistency in variables optimality, while thinner lines represent more variability in optimal design variables. Empty circles represent values that were not selected as optimal for any starting point. These patterns illustrate which design variables were consistently identified as optimal and benchmark each search configuration against the optimal solutions obtained from the full factorial search. The optimal design solutions derived from the initial run of the ungrouped sequential search significantly differ from those identified through the full factorial search. Additionally, varying starting points yield different optimal design solutions across experiments. Specifically, the ungrouped search across starting points identified the same primary trends as the full factorial search for 6 out of 11 variables (window-to-wall ratio, thermal mass, U-value of insulation, U-value of glazing, HVAC plant and distribution system type), compared to 5 out of 11 variables from the lower bound search (window-to-wall ratio, thermal mass, U-value of glazing, HVAC plant and distribution system type).

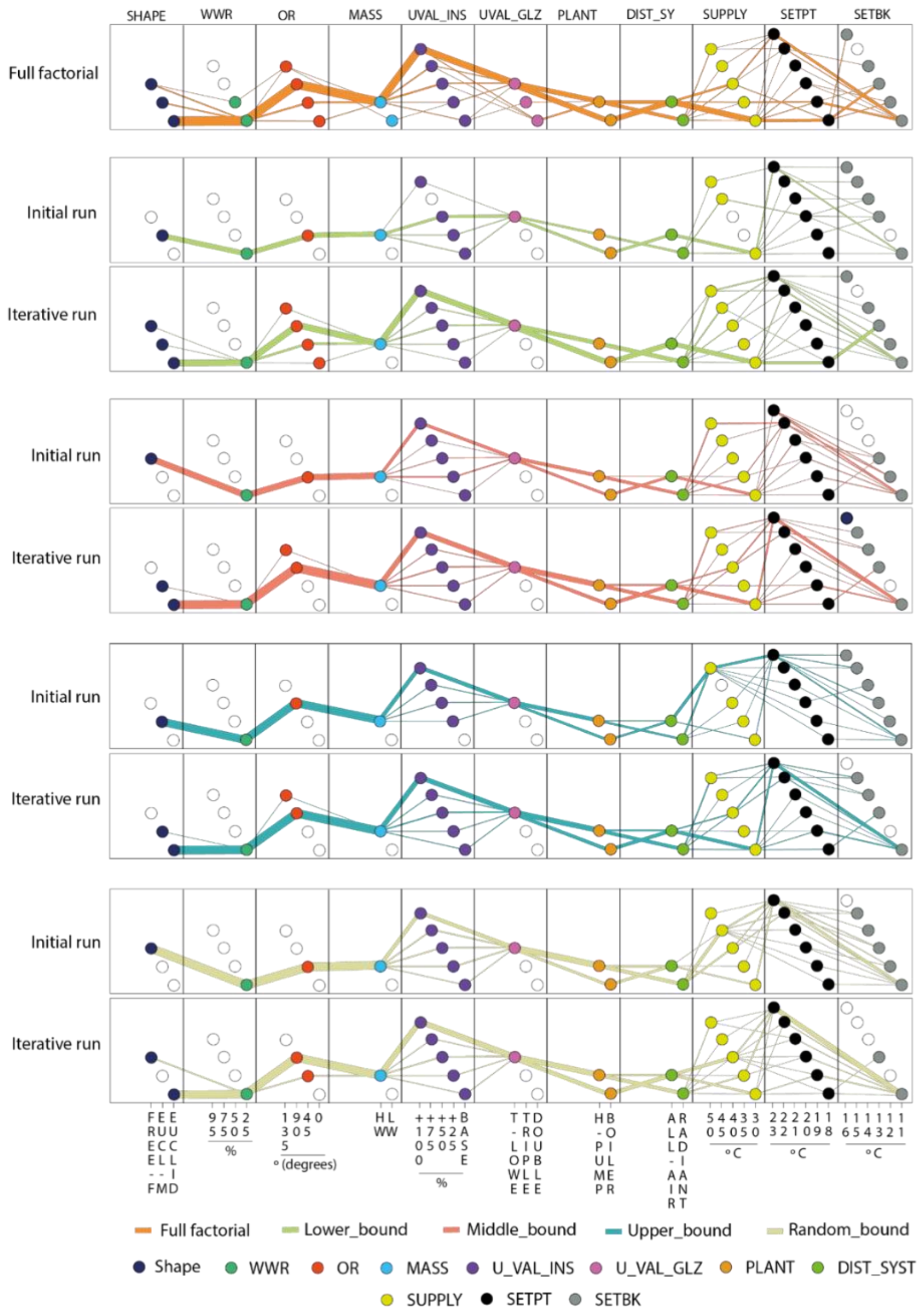

Figure B.1. Design variable trends of the initial and iterative ungrouped sequential search with different starting bounds (green, red, blue and yellow lines) and the full factorial (orange lines).

However, Figures B.2 and B.3 demonstrate that increasing the grouping of variables results in (1) the identification of a greater number of building design options that frequently appear in the full factorial search, and (2) a reduction in sensitivity to the initial starting points. In fact, the grouping of variables within the related design elements (e.g., geometry, fabric, HVAC systems) results in a higher consistency of optimal variable trends across starting points compared to the ungrouped approach, as seen in the thicker, more uniform lines for most parameters, and demonstrates reduced sensitivity to initial starting bounds, indicated by fewer deviations in line thickness. The element grouped search replicates the main trends of the full factorial search for 8 out of 11 variables (excluding supply water temperature, setpoint and setback temperatures). Finally, the field grouped search achieves the highest alignment with the full factorial approach, as shown by the thick, consistent lines across nearly all variables, reflecting uniform parameter selection regardless of starting point. This search configuration effectively captures the interactions among design variables, aligning with the global optima and underscores its ability to consistently identify key design variables with minimal sensitivity to initial conditions. The field grouped search captures 10 out of 11 variables (with only the setpoint temperature missing).

The iterative run (1) identifies the main trends in design options for experimental tests with a large number of different options during the initial run compared to the full factorial search (ungrouped search), (2) increases the number of Pareto optimal solutions aligning with the main trends from the initial run's configurations, and (3) identifies the patterns of design options related to secondary trends not captured during the initial run, thereby matching all design options obtained during the full factorial search. Despite the variations in Pareto optimal solutions compared to the full factorial search, all grouping experiments consistently identified the same high-impact design options regarding energy demand and thermal comfort. This indicates that, irrespective of variable grouping, the sequential search can identify all design options with high sensitivity to the optimization objectives, while it might miss design options with lower or negligible impact (Figure B.3).

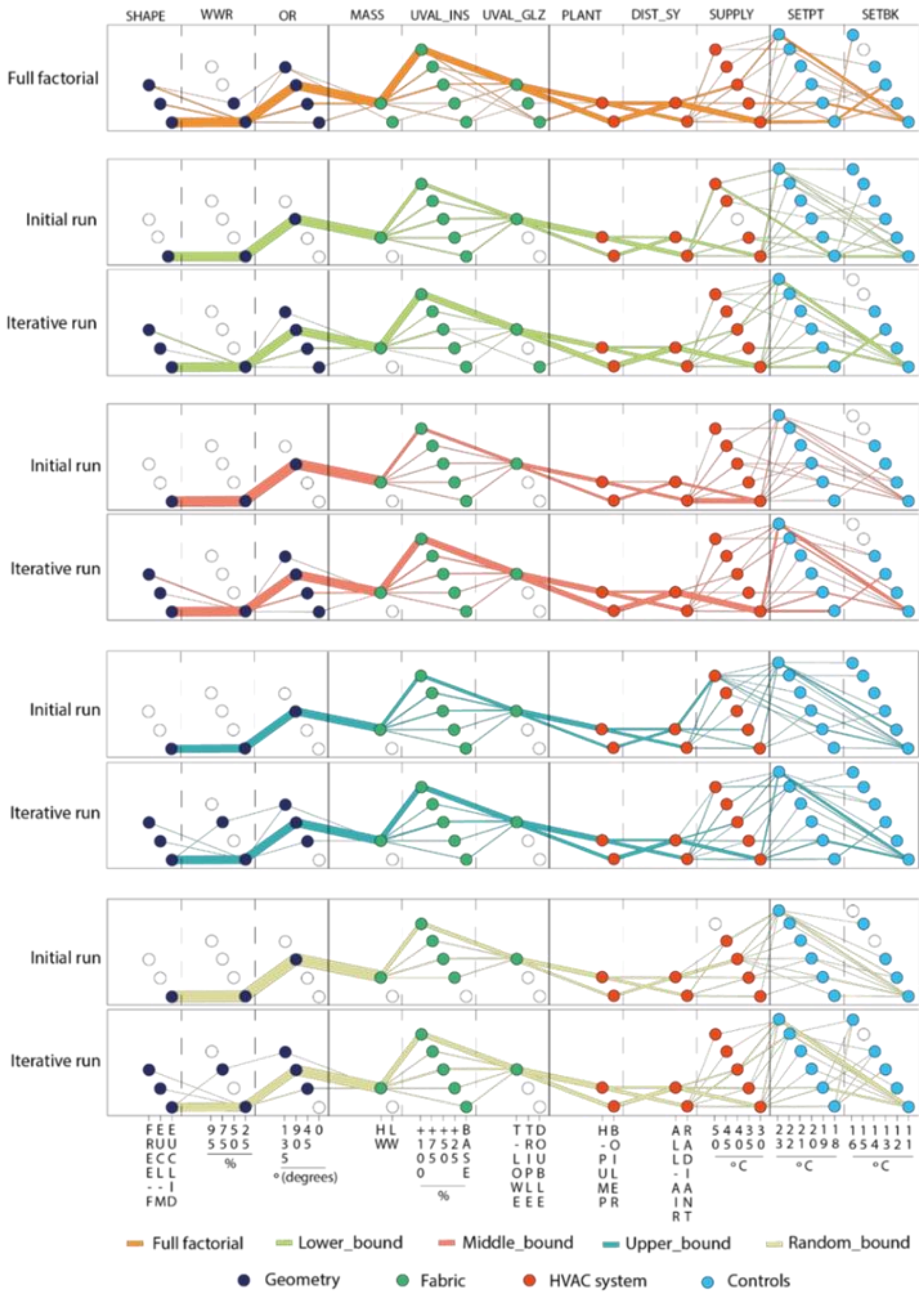

Figure B.2. Design variable trends of the initial and iterative element grouped sequential search with different starting bounds (green, red, blue and yellow lines) and the full factorial (orange lines).

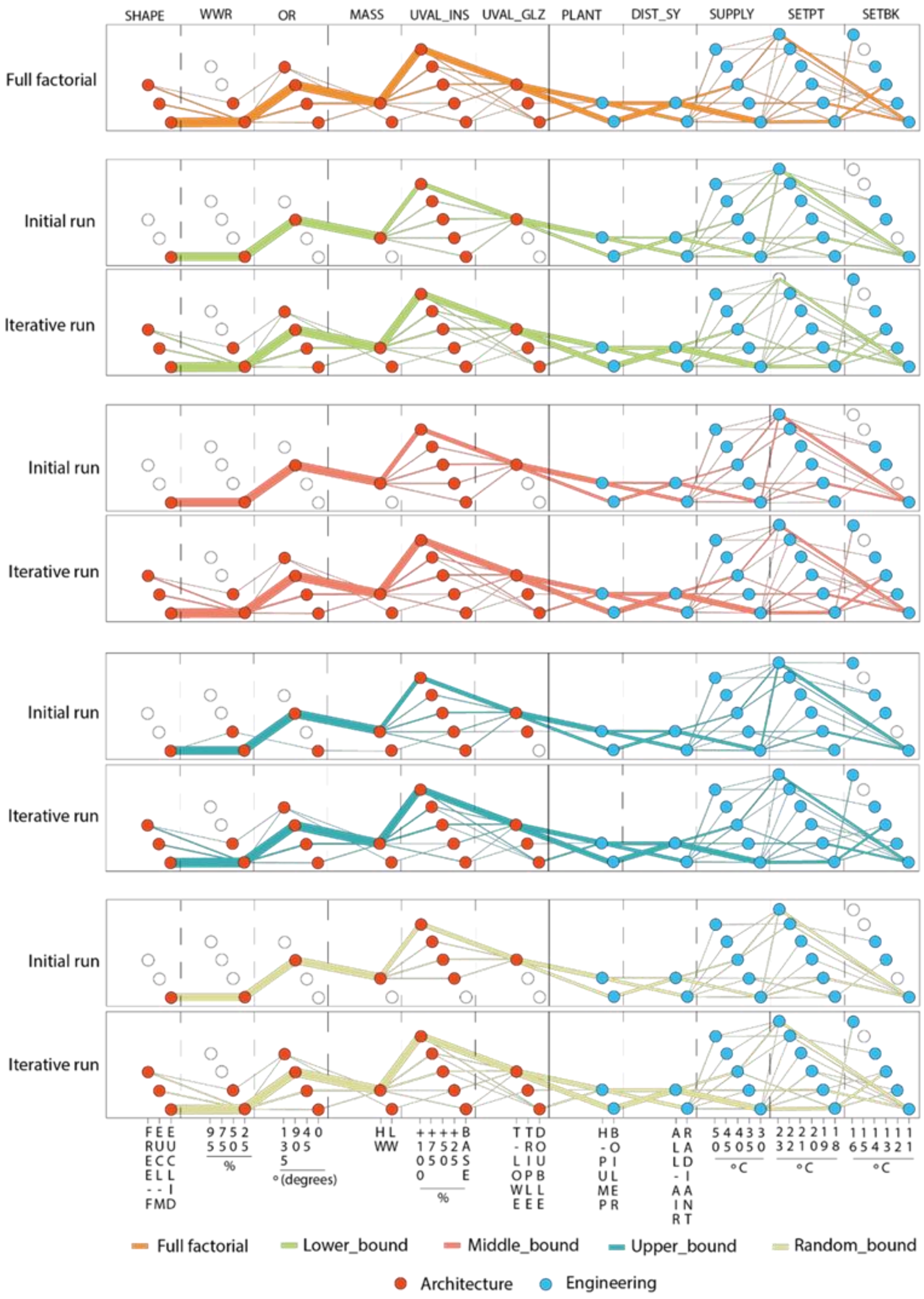

Figure B.3. Design variable trends of the initial and iterative field grouped sequential search with different starting bounds (green, red, blue and yellow lines) and the full factorial (orange lines).

## Sensitivity analysis of design variables

The impact of each design option on the optimization objectives is evaluated using sensitivity analysis to identify the most influential variables and ensure the search captures their dependencies. The enhanced Morris method, a reliable One-At-A-Time (OAT) screening technique combined with factorial sampling, is employed [37]. This method refines the sampling strategy to improve input space exploration for models with numerous input factors without increasing model evaluations, maximizing trajectory dispersion in the input factor space [38]. Widely applied in building energy analysis [39-41], the Morris method models n parameters, x = (x1,…,xn), as dimensionless and normalized, varying across r selected discretized trajectories (evenly distributed points) within an n-dimensional design space (hypercube), with model output y = f(x). It estimates the elementary effect (EE), representing the relative magnitude of change in each input parameter through an OAT sampling process on the output y. For each trajectory, a random starting point in the input space is generated, adjusting one factor at a time in random order. The EE for the ith input factor at a given value of x is defined as follows in Equation B.1:

$$EE_i(x) = \frac{[y(x_1,\dots,x_{i-1},x_i + \Delta, x_{i+1},\dots,x_n) - y(x)]}{\Delta} \quad \text{(B.1)}$$

Parameter sampling defines a region of exploration within the design space, where the elementary effect (EE) quantifies the impact of offsetting a parameter by a step size (Δ) at randomly selected points along a sequence (trajectory). Each point in the trajectory represents a single evaluation run, and the EE is calculated between two one-at-a-time (OAT) simulation runs, where one input factor is varied. The sensitivity of each design parameter is evaluated using two metrics: the mean of the absolute EE distribution (μ*) (Equation B.2), which ranks factors by their influence on the output, distinguishing influential from negligible parameters, and the standard deviation (σ) (Equation B.3), which captures higher-order effects due to interactions with other factors. A high σ indicates non-linear or non-monotonic behaviour.

$$\mu^* = \frac{1}{r}\sum_{i=1}^{r} |EE_i| \quad \text{(B.2)}$$

$$\sigma = \sqrt{\frac{1}{(r-1)}\sum_{i=1}^{r} |EE_i - \mu|^2} \quad \text{(B.3)}$$

Figure B.4 presents the results of the sensitivity analysis to assess the impact of each design variable on the optimization objectives. The ranking bar charts display the sensitivity of each design variable, while the scatterplot examines the relationship between standard deviation and mean for each variable on energy demand and thermal comfort. The design variables of which the sequential search always matched the full factorial search trends are the HVAC plant type, building fabric thermal mass, window-to-wall ratio, and HVAC distribution type and displayed the highest sensitivity on energy demand and thermal comfort. Moreover, the element and field grouped searches more consistently identified variables with lower sensitivity such as building shape, orientation, insulation thickness, windows U-values, supply water temperatures and setpoint and setback temperatures.

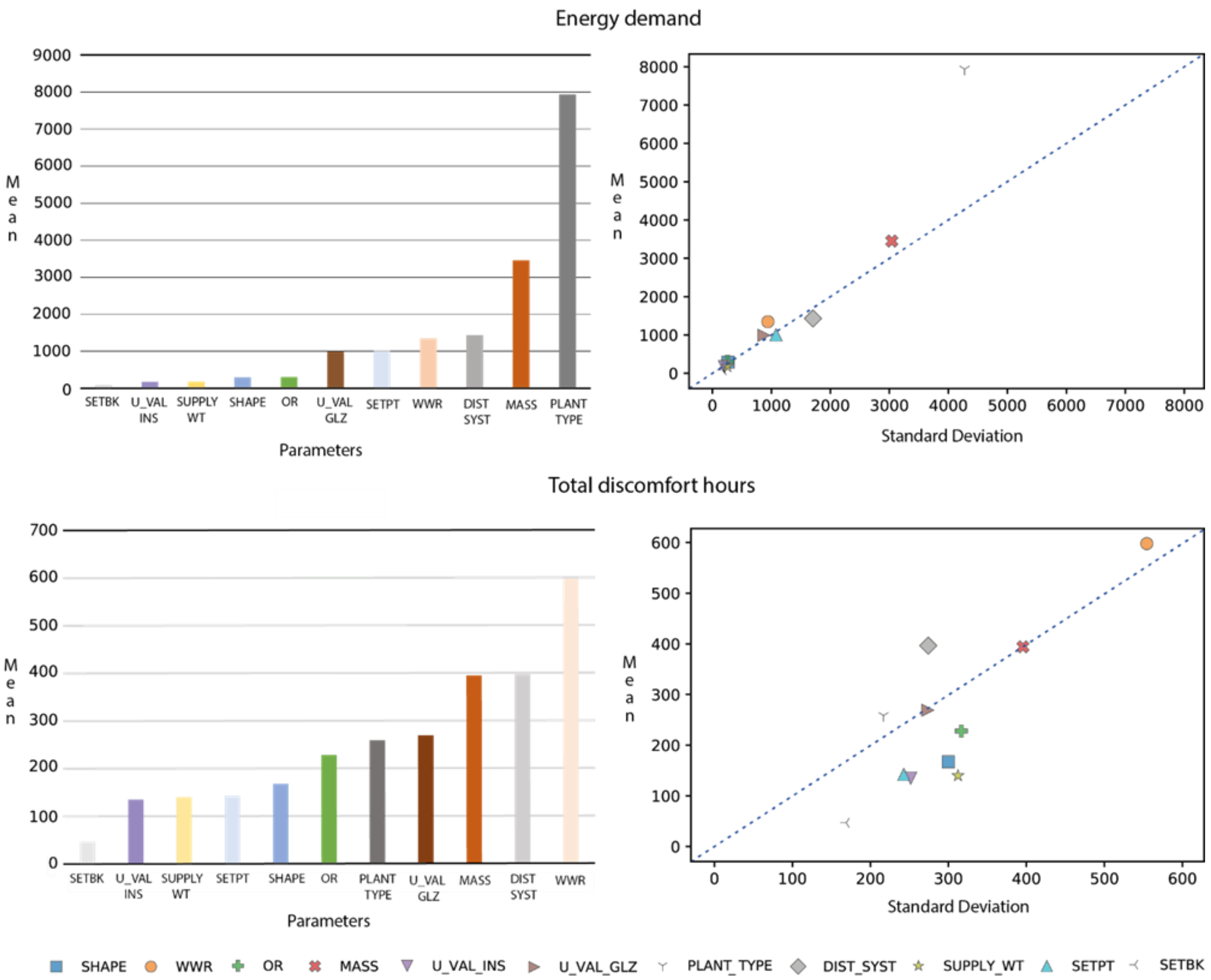

Figure B.4. Ranking bar charts displaying parameters' impact (left), and elementary effects scatterplots (right) on energy demand (above) and thermal comfort (below).